\documentclass{article}

\PassOptionsToPackage{numbers, compress}{natbib}
 \usepackage[preprint]{neurips_2026}

\usepackage[utf8]{inputenc} 
\usepackage[T1]{fontenc}    
\usepackage{hyperref}       
\usepackage{url}            
\usepackage{booktabs}       
\usepackage{amsfonts}       
\usepackage{nicefrac}       
\usepackage{microtype}      
\usepackage{xcolor}         
\usepackage{colortbl}
\usepackage{etoc}
\usepackage{graphicx}
\usepackage{subcaption}
\usepackage{algorithm}
\usepackage{algorithmic}
\usepackage{multirow}
\usepackage{multicol}
\usepackage[table]{xcolor}
\usepackage{arydshln}
\usepackage{amsmath}
\usepackage[capitalize]{cleveref}
\usepackage{wrapfig,lipsum,booktabs}
\crefname{section}{Section}{Sections}
\crefname{table}{Table}{Tables}
\crefname{figure}{Figure}{Figures}
\crefname{appendix}{Appendix}{Appendix}
\usepackage{amsmath} 
\usepackage{amssymb} 
\newcommand{\E}{\mathbb{E}}

\title{SAFE-SVD: Sensitivity-Aware Fidelity-Enforcing SVD for Physics Foundation Models}

\author{Chengjie Hong$^2$, Feixiang He$^3$, Yiheng Zeng$^2$, Lulu Kang$^4$, He Wang $^{1,2}$\thanks{Corresponding author, he\_wang@ucl.ac.uk}\\$^1$AI Centre, University College London, UK\ \ \ \ $^2$University College London, UK\\ $^3$Central South University, China\ \ \ \ $^4$ University of Massachusetts at Amherst, US}

\input{preamble}

\begin{document}

\maketitle

\begin{abstract}
\hypertarget{sec:abstract}{}
We propose a new method for compressing physics foundation models (PFMs) which is a new trend in AI for Science. While model compression is essential for reducing memory use and accelerating inference in large foundation models, it remains under-explored for PFMs, where preserving physical fidelity is crucial. The challenge lies in the functional nature of physics data, where partial derivatives encode spatiotemporal dynamics and exhibit high sensitivity to compression. Conventional compression methods ignore this structure, often causing severe performance degradation or failure. To address this, we introduce a sensitivity-aware fidelity-enforcing compression framework that explicitly models loss-aware layer sensitivity in the output function space during compression. This provides a new route to compressing scientific foundation models while preserving accuracy and physical fidelity. Experiments show substantial gains over existing methods across multiple models and datasets, achieving significantly higher compression ratios while maintaining accuracy, in some cases by orders of magnitude. More broadly, the work potentially leads to a new subfield of efficient, deployable, and sustainable scientific foundation models in AI for Science. The code is available at: \url{https://github.com/realcrane/SafeSVD}.
\end{abstract}

\section{Introduction}
\label{sec:introduction}
Physics foundation models (PFMs), analogous to foundational models in natural language processing, have emerged as general-purpose pre-trained models for a broad range of downstream scientific tasks, from simulation acceleration \cite{azizzadenesheli2024neural,rahman2024pretraining} to weather forecasting \cite{lam2023learning,nguyen2023climax}. 
Their success has sparked growing research interest \cite{subramanian2023towards,herde2024poseidon,soares2025towards}. 
However, as PFMs continue to scale, their increasing computational demands make fine-tuning and inference prohibitively expensive, limiting practical deployment. In parallel, model compression has proven effective for reducing the computational costs of large neural networks while preserving task performance~\cite{xu2023survey}. Yet despite rapid advances in compression for general foundation models, compression for PFMs remains largely unexplored.

Compressing PFMs presents unique challenges beyond those encountered in conventional foundation models. Unlike models trained on discrete data (\eg images or text), PFMs learn mappings over continuous function spaces defined over space and time, aiming not only to predict system behavior but also to capture governing dynamics and underlying physical laws. Consequently, compression must preserve not only predictive accuracy but also \textit{physical fidelity}, which, in our context, refers to preserving the derivative structures underlying PDE-specific physical properties, such as conservation laws, structural invariants, and higher-order derivative behavior. For example, in incompressible fluid modeling, accurate velocity prediction alone is insufficient if mass conservation is violated. Compression methods that neglect such constraints can produce physically implausible predictions or unstable long-term simulations. 

In this work, we focus on low-rank compression based on Singular Value Decomposition (SVD), a widely adopted technique due to its mathematical simplicity, hardware efficiency, and structured parameter reduction. While SVD-based compression has been extensively studied for large language models~\cite{achiam2023gpt} and more recently multimodal architectures~\cite{liu2024improved,kim2024openvla, yu2026elsa3d}, existing approaches are poorly suited to PFMs. A central limitation is that they optimize compression primarily around parameter reconstruction or task accuracy, without accounting for how layer-wise compression perturbations affect physical fidelity in the output function space. In particular, we observe that network layers contribute unequally to preserving physical structure, while their effects are strongly coupled through cross-layer error propagation. Existing methods either treat layer importance independently~\cite{hsu2022language}, compress layers separately~\cite{wang2025svdv1,wang2025svdv2}, or only partially model propagated compression errors without differentiating physics-sensitive layers~\cite{hu2026saes}. Moreover, these coupled dependencies make rank allocation especially challenging: manual tuning does not scale to large PFMs, while joint cross-layer optimization is highly non-convex~\cite{wang2025dobi}. As a result, existing methods may preserve strong zeroth-order numerical accuracy while violating fundamental physical laws.

To address these challenges, we propose a physics-aware SVD compression framework tailored for PFMs. Our key idea is to align compression with physical fidelity by analyzing the output function in Sobolev space, explicitly modeling sensitivity across multiple orders of partial derivatives with respect to network layers. Building on this, we perform sequential layer-wise compression while jointly accounting for intra-layer reconstruction quality and cross-layer error propagation. Guided by these sensitivity and dependency signals, we further introduce a greedy rank allocation strategy that adaptively computes desired compression ratio on different layers. We evaluate our method on multiple PFMs, including Poseidon~\cite{herde2024poseidon}, VICON~\cite{cao2024vicon}, and MPP~\cite{mccabe2024multiple}, across diverse datasets and physical systems, demonstrating consistent improvements in compression ratio, predictive performance, and physical fidelity. Our contributions are summarized as follows:
\begin{itemize}
    \item We propose, to the best of our knowledge, the first SVD-based physics-aware model compression framework for PFMs.
    \item We introduce a novel SVD-based method that jointly accounts for intra-layer reconstruction error, cross-layer perturbation propagation, and physics-informed layer importance.
    \item We develop a practical rank allocation strategy for scalable physics-aware compression.
    \item We conduct extensive experiments demonstrating significantly improved compression ratios, predictive accuracy, and physical fidelity across multiple PFMs and scientific domains.
\end{itemize}
\section{Related Work}
Existing methods can be divided into pruning, quantization, knowledge distillation, and low-rank approximation. Pruning removes redundant parameters, but either requires specialized hardware for acceleration or can incur substantial accuracy loss \cite{frantar2023sparsegpt,ashkboos2024slicegpt,ma2023llm,xia2024sheared, wang2026pare}. Quantization achieves strong memory savings, but depends on hardware support and optimized kernels, and aggressive low-bit settings often degrade accuracy \cite{dettmers2023qlora,lin2024awq,shang2023pb, wang2026pulsequant}. Knowledge distillation transfers knowledge from a large teacher model to a smaller student model, but typically requires costly retraining, which is expensive for billion-scale models \cite{zhao2022decoupled,wang2021knowledge}. Comparatively, low-rank approximation is an attractive hardware-agnostic post-training alternative.

SVD is a widely used low-rank approximation method for model compression. Some methods mainly focus on local reconstruction to minimize layer-wise output error, relying on \eg activation whitening~\cite{wang2025svdv1}, and input shift and error propagation modeling~\cite{hu2026saes}. However, these methods are local methods and are not directly linked to the global objective, \eg in our case task-specific or physics-fidelity losses. To this end, weight importance can be estimated w.r.t. the global objective~\cite{hsu2022language}, but with high computational costs while still ignoring cross-layer dependencies. Also, all the above methods are based on uniform rank selection. In contrast, other methods focus on non-uniform rank allocation to layers. They are based on heuristics~\cite{yuan2023asvd,wang2025svdv2} or joint rank optimization~\cite{wang2025dobi}, but both are suboptimal for highly sensitive layers. Some methods consider layer sensitivity during rank allocation~\cite{ding2025dipsvd}, but without linking layer sensitivity to the global objective, hence also a local method.
 
\section{Methodology}
\label{sec:methodology}
Our method is built on three key insights. First, to balance optimization tractability and compression quality, we adopt a sequential layer-wise compression scheme, where the compression of each layer is conditioned on the previously compressed layers while explicitly modeling the propagation of compression-induced errors across layers. Second, since layers contribute differently to the physical fidelity of the output, we introduce a physics-aware sensitivity metric which quantifies layer importance through its impact on the final loss, consisting of the original model loss augmented by an additional loss term defined in the Sobolev space. Unlike existing work~\cite{hsu2022language} which operates directly on model weights, inevitably leading to high computational cost for large models and insufficient modeling of layer sensitivity, our method operates on intermediate activations, enabling more efficient compression and effective sensitivity modeling. Third, under a global compression target, we propose a greedy rank allocation strategy which adaptively controls compression intensity across layers according to their estimated contribution to model performance.
\subsection{Problem Formulation}
Consider a pretrained model of $N$ linear layers with weights $\{\mathbf{W}_i\}_{i=1}^N$. Let $\{\mathbf{X}_i\}_{i=1}^N$ denote the corresponding intermediate activations, where $\mathbf{X}_1$ is the model input. $\mathbf{X}_i$ are random variables sampled from the data. After compression, the corresponding weights and activations become $\{\mathbf{W}_i^{\prime}\}_{i=1}^{N}$ and $\{\mathbf{X}_i^{\prime}\}_{i=1}^{N}$, respectively. 
Since joint optimization across all layers is difficult, we adopt a sequential layer-wise compression scheme which starts from the first layer and processes one layer at a time while propagating the effects of previously compressed layers. For a given layer, let the full-precision output be $\mathbf{Z} = \mathbf{W}\mathbf{X}$ omitting the subscripts for brevity. 
In deep networks, the input activation $\mathbf{X}$ itself is already perturbed by previous compression, so the actual compressed output becomes $\mathbf{Z}^{'} = \mathbf{W}'\mathbf{X}'$. We represent the resulting perturbation as $\Delta \mathbf{Z} = \mathbf{Z}^{'} - \mathbf{Z}$.

\paragraph{Loss-Aware Objective.} To preserve both predictive accuracy and physical fidelity, existing methods based on local reconstruction are insufficient~\cite{yuan2023asvd, wang2025svdv1, hu2026saes}. Therefore, we consider the effect of the intermediate perturbation $\Delta \mathbf{Z}$ on the final loss
\begin{equation}
\mathcal{L}
=
\mathcal{L}_{\mathrm{Model}}
+
\lambda_{\mathrm{Sob}}
\mathcal{L}_{\mathrm{Sob}},
\end{equation}
where $\lambda_{\mathrm{Sob}}$ controls the relative contribution of
the Sobolev loss $\mathcal{L}_{\mathrm{Sob}}$ with respect to the original model loss $\mathcal{L}_{\mathrm{Model}}$~\ref{app:model_base_loss}.
\begin{equation}
\mathcal{L}_{\mathrm{Sob}}^{(q)}
=
\sum_{|\nu|\le p}
\lambda_{\nu}
\frac{1}{|\Omega_h|}
\sum_{\mathbf{r}\in \Omega_h}
\left\|
D^{\nu}\mathbf{u}(\mathbf r)
-
D^{\nu}\hat{\mathbf{u}}(\mathbf r)
\right\|_{q}^{s},
\qquad q=s\in\{1,2\}.
\end{equation}
where $\mathbf{u}$ and $\hat{\mathbf{u}}$ denote the ground-truth and predicted fields on the discrete spatial grid $\Omega_h$, respectively, and $\mathbf{r}\in\Omega_h$ denotes a spatial grid point.
Here, $\nu$ denotes the derivative order, and $D^\nu$ denotes the differentiation operator used to approximate the corresponding derivatives, and $\lambda_\nu$ weights the associated derivative term. The parameter $q$ specifies the norm order, while $s$ denotes its exponent. We set $q=s\in\{1,2\}$, corresponding to the
$l_1$ Sobolev loss and the squared $l_2$ Sobolev loss, respectively. The parameter $p$ denotes the highest derivative order considered for each task; for example, incompressible Navier--Stokes uses derivatives up to the second order
(see Table~\ref{tab:pde_summary_fixed}). In \cite{li2021brecq}, the loss change $\Delta \mathcal{L} = \mathcal{L(\mathbf{W+\Delta \mathbf{W}})} - \mathcal{L(\mathbf{W})}$ is approximated by a Taylor series up to the second-order term. Then minimizing the expectation of $\Delta \mathcal{L}$ can be approximated by:
\begin{equation}
    \min_{\Delta \mathbf{W}} \,\, \mathbb{E}\left[ \Delta \mathcal{L}\right] \approx \min_{\Delta \mathbf{Z}} \,\, \frac{1}{2} \mathbb{E} \left[ \Delta \mathbf{Z}^\top \mathbf{F}_{\mathbf{z}}  \Delta\mathbf{Z} \right],
    \label{eq:objective_ori}
\end{equation}
where $\mathbf{F}_{\mathbf{z}} = \mathbf{g}\mathbf{g}^T$ and $\mathbf{g}=\nabla_{\mathbf{Z}} \mathcal{L}$. However, for PFMs, this would mean maintaining a per-sample $\mathbf{F}_{\mathbf{z}}$ for all data samples, which is prohibitively expensive in both memory and computation. Therefore, we approximate $\mathbf{F}_{\mathbf{z}}$ by its expectation $\mathbf{F}_{\mathbf{Z}} =  \mathbb{E} [\mathbf{g}\mathbf{g}^{\top}]$ over the whole calibration distribution, \ie the Fisher Information Matrix (FIM)~\ref{app:fisher_matrix_definition}. 
We refer to the sensitivity based on this approximation as Fisher-based sensitivity. In practice, $\mathbf{F}_{\mathbf{Z}}$ is symmetric positive definite \ref{app:pd}, then:
\begin{equation}
    \min_{\Delta \mathbf{Z}} \,\, \frac{1}{2} \mathbb{E} \left[ \Delta \mathbf{Z}^\top \mathbf{F}_{\mathbf{Z}}  \Delta\mathbf{Z} \right] = \min_{ \Delta \mathbf{Z}} \,\, \frac{1}{2} \mathbb{E} \left[ \| \mathbf{L} \Delta \mathbf{Z} \|_2^2 \right],
    \label{eq:objective}
\end{equation}
where $\mathbf{F}_{\mathbf{Z}} = \mathbf{L}^\top \mathbf{L}$ is its Cholesky decomposition. Then $\mathbf{L} \Delta \mathbf{Z}$ provides a principled measure of layer sensitivity with respect to the loss, and prioritizes perturbations that are most critical to preserving the physical fidelity included in the loss. 

Substituting $\Delta \mathbf{Z}$ into Equation \ref{eq:objective}, we obtain the layer-wise compression objective:
\begin{equation}
\label{eq:obj}
    \min_{ \mathbf{W}'} \,\, \frac{1}{2} \mathbb{E} \left[ \| \mathbf{L}(\mathbf{W}\mathbf{X} - \mathbf{W}'\mathbf{X}') \|_2^2 \right].
\end{equation}
However, we observe that optimizing this objective can be unstable and prone to overfitting to the calibration data. Empirically, we find that adding a regularization can improve stability and generalization. The regularization is an intra-layer reconstruction regularizer based on the clean input $\mathbf{X}$, which captures the reconstruction error of current compressing a layer, \ie $\mathbf{W}\mathbf{X} - \mathbf{W}'\mathbf{X}$. We further introduce a coefficient $\alpha\in(0, 1)$ to balance the intra-layer reconstruction error and the propagated error:
\begin{equation}
\label{eq:full_objective}
    \min_{ \mathbf{W}'} \,\, (1-\alpha) \underbrace{\mathbb{E} \left[ \| \mathbf{L}(\mathbf{W}\mathbf{X} - \mathbf{W}'\mathbf{X}) \|_2^2 \right]}_{\text{\color{red}{intra-layer reconstruction error}}} + \alpha \underbrace{\mathbb{E} \left[ \| \mathbf{L}(\mathbf{W}\mathbf{X} - \mathbf{W}'\mathbf{X}') \|_2^2 \right]}_{\text{\color{purple}{propagated error}}}.
\end{equation}
A similar strategy of combining the intra-layer reconstruction error and the propagated error has also been recently employed in \cite{hu2026saes}. However, the key difference is, while both of their errors are purely local to a layer, ours is global and measures loss-aware layer sensitivity by introducing $\mathbf{L}$. 
\paragraph{Calibration.} We assume access to a small calibration set $\mathcal{D}_{\mathrm{cal}}=\{(x_i, y_i)\}_{i=1}^{N_{\mathrm{cal}}}$ from the training data, where $(x_i, y_i)$ is an input-output pair.  These samples are only used to estimate activation statistics and compute the FIM to minimize Equation \ref{eq:full_objective} from the original model.


\subsection{Optimization via Low-Rank Approximation}
Although directly minimizing Equation \ref{eq:full_objective} subject to a pre-defined rank is possible, prior studies \cite{yuan2023asvd, wang2025svdv1} show that using SVD is an efficient solution, which requires to minimize the trace-equivalent form Equation \ref{eq:full_objective} using the identity $\|\mathbf{a}\|_2^2 = \operatorname{Tr}(\mathbf{a}\mathbf{a}^\top)$ (see Appendix \ref{app:Frobenius_to_trace}):
\begin{equation}
\label{eq:full_objective_trace_form}
    \min_{ \mathbf{W}'} \,\,  -2\operatorname{Tr}\left(\mathbf{L}\mathbf{W}'\boldsymbol{\Sigma}_{\text{cov}}\mathbf{W}^\top\mathbf{L}^\top\right) + \operatorname{Tr}\left(\mathbf{L}\mathbf{W}'\mathbf{\Sigma}_{\text{c\_cov}}\mathbf{W}'^\top\mathbf{L}^\top\right) + \text{const}, 
\end{equation}
where 
\begin{equation}
    \mathbf{\Sigma}_{\text{cov}}= (1-\alpha)\boldsymbol{\Sigma}_{\mathbf{X}\mathbf{X}} + \alpha\boldsymbol{\Sigma}_{\mathbf{X}'\mathbf{X}},\quad \mathbf{\Sigma}_{\text{c\_cov}}= (1-\alpha)\boldsymbol{\Sigma}_{\mathbf{X}\mathbf{X}} + \alpha\boldsymbol{\Sigma}_{\mathbf{X}'\mathbf{X}'}
    \label{eq:covariances}
\end{equation}
In practice, $\mathbf{\Sigma}_{\text{c\_cov}}$ is positive definite (see Appendix~\ref{app:pd}), allowing a Cholesky decomposition $\boldsymbol{\Sigma}_{\text{c\_cov}} = \mathbf{R}\mathbf{R}^\top$. Substituting this factorization into Equation \ref{eq:full_objective_trace_form}, this problem can be rewritten as a Frobenius-norm regression task (see Appendix \ref{app:trace_to_Frobenius}):
\begin{equation}
\label{eq:Frobenius_norm_regression}
     \min_{ \mathbf{W}'} \,\,\left\| \mathbf{L}\mathbf{W}'\mathbf{R} - \mathbf{L}\mathbf{W}\mathbf{\Sigma}^\top_{\text{cov}}\mathbf{R}^{-\top} \right\|_F^2,
\end{equation}
where if we define $\mathbf{M} = \mathbf{L}\mathbf{W}'\mathbf{R}$ and $\mathbf{M}^* = \mathbf{L}\mathbf{W}\mathbf{\Sigma}^\top_{\text{cov}}\mathbf{R}^{-\top}$,
then the optimization problem reduces to a standard low-rank approximation:
\begin{equation}
    \min_{\mathbf{M}} \ \|\mathbf{M} - \mathbf{M}^*\|_F^2 \quad \text{subject to } \operatorname{rank}(\mathbf{M}) = k.
\label{eq:LRObjective}
\end{equation}
By the Eckart--Young--Mirsky theorem, the optimal rank-$k$ solution is obtained by truncating $\mathbf{M}^*$ and mapping it back to the original parameter space and obtaining the compressed weight matrix:
\begin{equation}
    \mathbf{M} = \mathrm{SVD}_k(\mathbf{M}^*), \quad \mathbf{W}' = \mathbf{L}^{-1}\mathbf{M}\mathbf{R}^{-1}
\label{eq:15}.
\end{equation}

\subsection{Loss-aware Rank Allocation}
Optimizing Equation \ref{eq:LRObjective} essentially comes down to choosing $k$ for each layer, which is a combinatorial optimization problem \cite{zhang2015accelerating}. In practice, existing research often resorts to heuristics which are compatible with their specific optimization formulations~\cite{yuan2023asvd, wang2025svdv2, wang2025dobi}. Similarly, we propose a greedy algorithm to allocate ranks to each layer, subject to a pre-defined retention ratio $\eta$, which denotes the target parameter retention ratio for the linear-layer weights selected for compression, so the compression ratio is $1-\eta$. A lower retention ratio corresponds to a higher compression level.

In the sequential compression setting, directly optimizing the rank of $\mathbf{M}^*$ is intractable because $\mathbf{M}^* = \mathbf{L}\mathbf{W}\boldsymbol{\Sigma}^{\top}_{\text{cov}}\mathbf{R}^{-\top}$ depends on perturbed activations $\mathbf{X}'$, which are unavailable before compression. To circumvent this, we factorize the clean covariance $\boldsymbol{\Sigma}_{\mathbf{XX}} = \mathbf{R}^c\mathbf{R}^{c\top}$ and observe that when the perturbation is small, \ie $\mathbf{X}' \approx \mathbf{X}$, we have $\mathbf{M}^* \approx \mathbf{L}\mathbf{W}\mathbf{R}^c$, so the singular values of $\mathbf{L}\mathbf{W}\mathbf{R}^c$ serve as a reliable proxy for rank selection. Even under low retention ratios where the perturbation is no longer negligible, we empirically find that it remains an effective proxy for rank selection.

\paragraph{Greedy Global Rank Allocation.}
Building on the SVD described above, we formulate the rank selection task as a global resource allocation problem across all $N$ linear layers. For each layer $l$ with its weight matrix $\mathbf{W}_l \in \mathbb{R}^{d_{\text{out}}^{(l)} \times d_{\text{in}}^{(l)}}$, its rank-$k_l$ approximation is factorized as
\begin{equation}
    \mathbf{W}_l \approx \mathbf{W}_{l, k_l} = (\mathbf{U}_l\mathbf{\Sigma}_{l, k_l}^{1/2})(\mathbf{\Sigma}_{l, k_l}^{1/2}\mathbf{V}_l^{\top}).
\end{equation}
Increasing the rank by one corresponds to adding one singular component to $\mathbf{W}_{l, k_l}$, \ie one column to $\mathbf{U}_l\mathbf{\Sigma}_{l, k_l}^{1/2}$ and one row to $\mathbf{\Sigma}_{l, k_l}^{1/2}\mathbf{V}_l^{\top}$. 
The resulting cost is therefore $P^{(l)} = d_{\text{out}}^{(l)} + d_{\text{in}}^{(l)}$. To satisfy a target retention ratio $\eta \in (0, 1]$, we impose the global budget
\begin{equation}
    B = \eta \cdot \sum_{l=1}^{N} \left( d_{\text{out}}^{(l)} \times d_{\text{in}}^{(l)} \right),
\end{equation}
where $d_{\text{out}}^{(l)} \times d_{\text{in}}^{(l)}$ is the number of parameters in linear layer $l$ and $\sum_{l=1}^{N} \left( d_{\text{out}}^{(l)} \times d_{\text{in}}^{(l)} \right)$ is the total number of parameters in the linear-layer weight matrices selected for compression. So $B$ is the total number of parameters that need to be retained. Under this budget, our rank allocation method decides which singular components should be retained across all layers. 

For each layer, we evaluate every candidate singular component from the matrix $\mathbf{L}_l\mathbf{W}_l\mathbf{R^c}_l$,  whose $i$-th singular value is denoted as $\sigma_{l,i}$. 
Under the second-order Taylor approximation, retaining this component yields an estimated reduction in loss degradation proportional to $\sigma_{l,i}^2$, see Appendix \ref{appendix: loss degradation}. Since a retained component at layer $l$ incurs the cost $P^{(l)}$, we define its benefit-cost ratio as $v_{l,i} = \sigma_{l,i}^2 / P^{(l)}$. We then rank all candidate components from all layers by $v_{l,i}$ and greedily retain them in descending order until no remaining candidate can be selected without exceeding the budget.

The overall allocation is detailed in \cref{alg:rank_selection}: (i) \textbf{SVD.} Compute the squared singular values $\{\sigma_{l,i}^2\}$ of $\mathbf{L}_l\mathbf{W}_l\mathbf{R^c}_l$. (ii) \textbf{Scoring.} Evaluate the score $v_{l,i}$ for all candidate rank-1 components in all layers. (iii) \textbf{Ranking.} Merge all candidates and sort them by $v_{l,i}$ in descending order. (iv) \textbf{Selection.} Iteratively select the highest-scoring component, increase the corresponding $k_l$ by one, and subtract $P_l$ from the remaining budget until the budget is exhausted. This yields the global rank configuration $\{k_1, \dots, k_N\}$, which is then used to compute the optimal compressed weight matrix $\mathbf{W}'_l$ according to Equation \ref{eq:15}.

\begin{algorithm}
\caption{Greedy Global Rank Allocation}
\label{alg:rank_selection}
\begin{algorithmic}[1]
\REQUIRE Budget $B$, Layer matrices $\{L_l W_l R^c_l\}_{l=1}^N$, Unit costs $\{P^{(l)}\}_{l=1}^N$
\ENSURE Rank configuration $\{k_l\}_{l=1}^N$

\STATE Initialize $k_l \leftarrow 0$ for all $l \in \{1, \dots, N\}$, and $\text{cost} \leftarrow 0$
\STATE $\mathcal{C} \leftarrow \bigcup_{l=1}^N \left\{ \left( \frac{\sigma_{l,i}^2}{P_l}, l \right) \mid \sigma_{l,i} \in \text{SVD}(L_l W_l R^c_l) \right\}$ \COMMENT{Global efficiency pool}
\STATE Sort $\mathcal{C}$ in descending order by efficiency score $v = \sigma^2 / P$

\FOR{each $(v, l) \in \mathcal{C}$}
    \IF{$\text{cost} + P^{(l)} \leq B$}
        \STATE $k_l \leftarrow k_l + 1$
        \STATE $\text{cost} \leftarrow \text{cost} + P^{(l)}$
    \ELSE
        \STATE \textbf{continue} \COMMENT{Skip this component because it would exceed the budget}
    \ENDIF
\ENDFOR

\RETURN $\{k_1, k_2, \dots, k_N\}$
\end{algorithmic}
\end{algorithm}

\section{Experiments}
\label{sec:experiments}
We evaluate our method on three representative PFMs: Poseidon~\cite{herde2024poseidon} VICON~\cite{cao2024vicon} and MPP~\cite{mccabe2024multiple}. We only show partial results on Poseidon and VICON in the main paper due to space limit and give additional results in Appendix~\ref{app:additional_exp_res}. For Poseidon, we fine-tune the pretrained model on three PDE families: incompressible Navier–Stokes, compressible Euler, and acoustic wave equations. We report one representative benchmark from each family, namely NS-PwC, CE-RM, and Wave-Gauss. In contrast, VICON is a unified model designed for multiple domains, and is evaluated separately on each domain. Calibration data are sampled from the corresponding training sets. For Poseidon, we construct the calibration set from 64 trajectories, using 210 samples per trajectory for Navier–Stokes and compressible Euler tasks, and 120 samples per trajectory for wave tasks. For VICON, we construct a fused calibration dataset from its supported domains \cite{takamoto2022pdebench}. Specifically, we use 2,048 calibration samples per scenario for VICON. All experiments are conducted on NVIDIA H200 GPUs. More details are presented in the Appendix \ref{app:dataset_config}.
\paragraph{Baselines} 
We compare our method with five representative SVD-based compression baselines which cover the major design dimensions of low-rank model compression, such as rank selection and weight reconstruction techniques. To ensure a fair comparison and focus on the effectiveness of compression, all methods are evaluated post-training without any additional fine-tuning. \textbf{FWSVD}~\cite{hsu2022language} performs Fisher-based layer-wise compression. \textbf{ASVD}~\cite{yuan2023asvd} applies activation-aware scaling before SVD decomposition. \textbf{SVD-LLM V2}~\cite{wang2025svdv2} improves layer-wise calibration and reconstruction for large models. \textbf{Dobi-SVD}~\cite{wang2025dobi} performs joint cross-layer rank optimization over all layers simultaneously. \textbf{SAES-SVD}~\cite{hu2026saes} further accounts for both intra-layer reconstruction error and inter-layer propagation error. Together, these methods span Fisher-based, activation-aware, globally optimized, and error-propagated compression strategies, providing strong baselines for comparison.

\paragraph{Metrics.}
We evaluate compression performance using two categories of metrics. General metrics measure overall predictive accuracy and functional smoothness, including the model’s original loss (\eg relative $l_1$ or MSE, depending on the backbone, and the Sobolev loss (physically meaningful derivatives depending on the PDE class, \eg second-order for Navier--Stokes equations), which captures consistency in both field values and derivative structures between the prediction and ground-truth. We additionally report PDE-specific metrics that capture task-dependent physical constraints. For incompressible Navier–Stokes, we evaluate divergence-free error and vorticity accuracy to assess incompressibility and rotational dynamics. To ensure fair comparison across models, datasets, and metrics with different numerical scales, we report the absolute relative change of each metric \(m\) from the uncompressed model (Origin):
\begin{equation}
\delta_m =
\frac{|m_{\mathrm{comp}}-m_{\mathrm{orig}}|}
{|m_{\mathrm{orig}}|+\epsilon}
\times 100\%,
\end{equation}
where \(m_{\mathrm{comp}}\) and \(m_{\mathrm{orig}}\) are the metric values of the compressed and uncompressed models, respectively, and \(\epsilon>0\) prevents division by zero. Lower percentages indicate closer agreement with Origin.
Detailed definitions of all metrics are provided in the Appendix \ref{app:detailed_metric}. 

\subsection{Ablation Study}\
\label{main:abalation_study}
\begin{table}[tb]
\vspace{-1em}
\centering
\caption{Ablation study on NS-BB at target parameter retention ratio
$\eta=0.2$. RS: rank selection; SE: sensitivity estimation; SL Type:
Sobolev loss type; SL Order: Sobolev order; $\alpha$: trade-off coefficient.}
\label{tab:ablation}
\scriptsize
\setlength{\tabcolsep}{2.2pt}
\renewcommand{\arraystretch}{0.95}
\resizebox{0.85\linewidth}{!}{%
\begin{tabular}{ccccc c cccc}
\toprule
\multicolumn{5}{c}{\textbf{Components}} 
& \multirow{2}{*}{\textbf{Size}} 
& \multirow{2}{*}{\textbf{Base Loss}$\downarrow$} 
& \multirow{2}{*}{\textbf{Sobolev}$\downarrow$} 
& \multirow{2}{*}{\textbf{Div-free}$\downarrow$} 
& \multirow{2}{*}{\textbf{Vorticity}$\downarrow$} \\
\cmidrule(r){1-5}
\textbf{RS} & \textbf{SE} & \textbf{SL Type} & \textbf{SL Order} & $\boldsymbol{\alpha}$ 
&  &  &  &  &  \\
\midrule
$\times$ & $\times$ & -   & - & -   & 125.9M & 2809\%  & 5018\%  & 2863\%  & 1971\% \\
$\checkmark$ & $\times$ & -   & - & -   & 126.3M & 63.92\% & 361.9\% & 126.1\% & 60.23\% \\
$\checkmark$ & $\checkmark$ & $l_2$ & 0 & 0.5 & 126.3M & 36.69\% & 168.8\% & 49.50\% & 19.97\% \\
$\checkmark$ & $\checkmark$ & $l_2$ & 1 & 0.5 & 126.3M & 37.52\% & 162.6\% & 45.53\% & 18.50\% \\
$\checkmark$ & $\checkmark$ & $l_2$ & 2 & 0.5 & 126.3M & 34.35\% & 152.2\% & 41.89\% & 17.33\% \\
$\checkmark$ & $\checkmark$ & $l_1$  & 0 & 0.5 & 126.3M & 19.72\% & 95.76\% & 23.84\% & 11.84\% \\
$\checkmark$ & $\checkmark$ & $l_1$  & 1 & 0.5 & 126.3M & 2.85\%  & 7.23\%  & 1.15\%  & 1.17\% \\
$\checkmark$ & $\checkmark$ & $l_1$  & 2 & 0.5 & 126.3M & 1.22\%  & 0.66\%  & 0.26\%  & 0.74\% \\
\midrule
$\checkmark$ & $\checkmark$ & $l_1$  & 2 & 0.3 & 126.3M & 1.32\%  & 0.59\%  & 0.24\%  & 0.75\% \\
$\checkmark$ & $\checkmark$ & $l_1$  & 2 & 0.7 & 126.3M & 1.20\%  & 0.83\%  & 0.29\%  & 0.77\% \\
\bottomrule
\end{tabular}
}
\vspace{-1em}
\end{table}

We separately evaluate our layer dependency modeling, layer sensitivity estimation (SE), and rank selection (RS). To test layer dependency, we use a base model without the Sobolev loss, loss-aware layer sensitivity, and rank selection. Then we add rank selection. Next, for SE, we add the Sobolev loss and loss-aware layer sensitivity. The Sobolev term includes derivatives of orders 0/1/2 to capture the principal physical structures, and is evaluated using both $l_2$ and $l_1$ norms. We evaluate different variants on Poseidon fine-tuned on NS-BB at a target retention ratio $\eta=0.2$, as shown in Table \ref{tab:ablation}. In our result tables, the Base Loss, Sobolev, Div-Free, Vorticity columns report the absolute relative changes of model's original loss, Sobolev loss, and the task-dependent physical constraints, respectively.
The Size column reports the total number of model parameters after
compression, including parameters that remain uncompressed.

Our base model implemented as $\E[||\mathbf{WX} - \mathbf{W'X}||_2^2 + ||\mathbf{WX-W'X'}||_2^2]$ with a uniform parameter retention ratio across layers (Appendix~\ref{app:uniform_rank}) and without layer sensitivity estimation (SE), resulting in severe degradation, with errors increasing by ranging from 1971\% to 5018\%. Replacing uniform allocation with the proposed RS substantially improves performance: reducing relative increase to $63.92\%$ for Base Loss, $361.9\%$ for Sobolev Loss, $126.1\%$ for Divergence, and $60.23\%$ for Vorticity, demonstrating the effectiveness of our global rank selection. Incorporating SE further improves performance. While the formulation is naturally aligned with the $l_2$ norm, our Poseidon ablation shows that the $l_1$ variant performs better in this setting, and we therefore adopt it as the default. Under Order-0 with SE and $l_1$, the errors are reduced to $19.72\%$, $95.76\%$, $23.84\%$, and $11.84\%$, respectively. Increasing the order consistently improves performance: Order-1 reduces errors to $2.85\%$, $7.23\%$, $1.15\%$, and $1.17\%$, while Order-2 achieves the best overall results with only $1.22\%$, $0.66\%$, $0.26\%$, and $0.74\%$ degradation. These trends indicate that higher-order Sobolev losses lead to more accurate layer sensitivity estimation and better preservation of physical structures of derivatives. Varying $\alpha$ further demonstrates the importance of layer dependency modeling. The balanced setting $\alpha=0.7$ achieves the lowest Base Loss while remaining competitive on physics-related metrics. This suggests that jointly balancing local reconstruction and cross-layer propagation is critical. Overall, the ablation confirms that loss-aware layer sensitivity, cross-layer error propagation, and global rank selection are all necessary for high-fidelity compression of PFMs.

\begin{table*}[tb]
\centering
\caption{Experimental Results on Poseidon.}
\label{tab:poseidon}
\resizebox{\textwidth}{!}{
\begin{tabular}{c c c r r r r r r r r}
\toprule
\multirow{2}{*}{\textbf{$\eta$}} &
\multirow{2}{*}{\textbf{Method}} &
\multirow{2}{*}{\textbf{Size}} &
\multicolumn{4}{c}{\textbf{NS-PwC}} &
\multicolumn{2}{c}{\textbf{CE-RM}} &
\multicolumn{2}{c}{\textbf{Wave-Gauss}} \\
\cmidrule(lr){4-7}
\cmidrule(lr){8-9}
\cmidrule(lr){10-11}

& & &
\textbf{Base Loss $\downarrow$} &
\textbf{Sobolev $\downarrow$} &
\textbf{Div-Free $\downarrow$} &
\textbf{Vorticity $\downarrow$} &
\textbf{Base Loss $\downarrow$} &
\textbf{Sobolev $\downarrow$} &
\textbf{Base Loss $\downarrow$} &
\textbf{Sobolev $\downarrow$} \\
\midrule
1 & Origin & 628.6M & 0.00\% & 0.00\% & 0.00\% & 0.00\% & 0.00\% & 0.00\% & 0.00\% & 0.00\% \\
\midrule
\multirow{6}{*}{0.6} & ASVD & 376.9M & 4496\% & 2680\% & 2017\% & 2805\% & 25.39\% & 6.61\% & 352.1\% & 6202\% \\
 & FWSVD & 377.1M & 7124\% & 2977\% & 1700\% & 3727\% & 23.24\% & 8.51\% & 569.7\% & 2415\% \\
 & SVD-LLM V2 & 438.3M & 7038\% & 2449\% & 1482\% & 3651\% & 26.78\% & 9.31\% & 576.4\% & 4982\% \\
 & Dobi-SVD & 378.1M & 7468\% & 3036\% & 2137\% & 3719\% & 35.70\% & 24.58\% & 658.6\% & 9210\% \\
 & SAES-SVD & 377.1M & 2882\% & 1763\% & 1572\% & 1863\% & 2.19\% & 2.53\% & 225.8\% & 3354\% \\
\hdashline
\rowcolor{gray!15} & \textbf{Ours} & 377.5M & \textbf{0.00\%} & \textbf{0.01\%} & \textbf{0.00\%} & \textbf{0.01\%} & \textbf{0.00\%} & \textbf{0.00\%} & \textbf{0.00\%} & \textbf{0.01\%} \\
\midrule
\multirow{6}{*}{0.4} & ASVD & 251.4M & 5169\% & 4364\% & 2773\% & 3311\% & 34.10\% & 15.79\% & 475.4\% & 14572\% \\
 & FWSVD & 251.5M & 7846\% & 5080\% & 2449\% & 3961\% & 28.33\% & 28.64\% & 588.1\% & 3542\% \\
 & SVD-LLM V2 & 374.9M & 8741\% & 6051\% & 4530\% & 4131\% & 35.87\% & 54.99\% & 656.0\% & 19627\% \\
 & Dobi-SVD & 252.6M & 8481\% & 5206\% & 2709\% & 4048\% & 39.27\% & 25.91\% & 670.4\% & 10497\% \\
 & SAES-SVD & 251.5M & 3062\% & 1868\% & 1632\% & 1967\% & 3.46\% & 6.18\% & 249.3\% & 4020\% \\
\hdashline
\rowcolor{gray!15} & \textbf{Ours} & 251.9M & \textbf{0.26\%} & \textbf{0.08\%} & \textbf{0.03\%} & \textbf{0.17\%} & \textbf{0.01\%} & \textbf{0.05\%} & \textbf{0.04\%} & \textbf{0.40\%} \\
\midrule
\multirow{6}{*}{0.2} & ASVD & 125.8M & 6725\% & 6183\% & 3563\% & 4167\% & 41.02\% & 80.05\% & 525.0\% & 13007\% \\
 & FWSVD & 125.9M & 8622\% & 7650\% & 3770\% & 4276\% & 40.57\% & 97.34\% & 655.1\% & 11452\% \\
 & SVD-LLM V2 & 353.2M & 9529\% & 6011\% & 4183\% & 4434\% & 61.44\% & 66.29\% & 1122\% & 19865\% \\
 & Dobi-SVD & 127.0M & 8774\% & 10090\% & 4999\% & 4440\% & 44.78\% & 29.72\% & 728.6\% & 20124\% \\
 & SAES-SVD & 125.9M & 3726\% & 2429\% & 2199\% & 2426\% & 6.25\% & 9.83\% & 366.9\% & 5873\% \\
\hdashline
\rowcolor{gray!15} & \textbf{Ours} & 126.3M & \textbf{2.64\%} & \textbf{1.79\%} & \textbf{0.52\%} & \textbf{1.35\%} & \textbf{0.10\%} & \textbf{1.01\%} & \textbf{0.70\%} & \textbf{151.9\%} \\
\bottomrule
\end{tabular}
}
\end{table*}

\begin{table*}[t]
\centering
\caption{Experimental Results on VICON}
\label{tab:vicon}
\resizebox{\textwidth}{!}{
\begin{tabular}{c c c r r r r r r r r}
\toprule
\multirow{2}{*}{\textbf{$\eta$}} &
\multirow{2}{*}{\textbf{Method}} &
\multirow{2}{*}{\textbf{Size}} &
\multicolumn{4}{c}{\textbf{NS-2D}} &
\multicolumn{2}{c}{\textbf{Comp-2D}} &
\multicolumn{2}{c}{\textbf{Euler-2D}} \\
\cmidrule(lr){4-7}
\cmidrule(lr){8-9}
\cmidrule(lr){10-11}

& & &
\textbf{Base Loss $\downarrow$} &
\textbf{Sobolev $\downarrow$} &
\textbf{Div-Free $\downarrow$} &
\textbf{Vorticity $\downarrow$} &
\textbf{Base Loss $\downarrow$} &
\textbf{Sobolev $\downarrow$} &
\textbf{Base Loss $\downarrow$} &
\textbf{Sobolev $\downarrow$} \\
\midrule
1 & Origin & 87.8M & 0.00\% & 0.00\% & 0.00\% & 0.00\% & 0.00\% & 0.00\% & 0.00\% & 0.00\% \\
\midrule
\multirow{7}{*}{0.6} & ASVD & 52.4M & 10140\% & 528.7\% & 7.01\% & 476.1\% & 41318\% & 1394\% & 2692\% & 97.36\% \\
 & FWSVD & 52.7M & 1861\% & 302.3\% & 2.23\% & 357.5\% & 7885\% & 947.6\% & 805.1\% & 57.72\% \\
 & SVD-LLM V2 & 52.7M & 73.76\% & 11.74\% & 5.30\% & 43.87\% & 260.0\% & 60.00\% & 47.26\% & \textbf{3.31\%} \\
 & Dobi-SVD & 52.9M & 1968\% & 109.3\% & 24.08\% & 163.6\% & 757.4\% & 260.1\% & 128.6\% & 18.49\% \\
 & SAES-SVD & 52.7M & 49.63\% & 7.06\% & 3.92\% & 26.01\% & 144.7\% & 33.70\% & 36.66\% & 25.99\% \\
\hdashline
 & \textbf{Ours} & 52.8M & 2.07\% & 1.22\% & \textbf{0.21\%} & 3.08\% & \textbf{16.54\%} & \textbf{3.82\%} & \textbf{3.08\%} & 28.35\% \\
 \rowcolor{gray!15} & \textbf{Ours*} & 52.8M & \textbf{0.51\%} & \textbf{0.02\%} & 0.71\% & \textbf{0.06\%} & 18.05\% & 10.86\% & 8.52\% & 28.89\% \\
\midrule
\multirow{7}{*}{0.4} & ASVD & 33.5M & 13064\% & 532.5\% & 11.34\% & 490.4\% & 46290\% & 1382\% & 2865\% & 94.70\% \\
 & FWSVD & 35.2M & 4862\% & 349.7\% & 3.24\% & 437.5\% & 22440\% & 1155\% & 2010\% & 69.01\% \\
 & SVD-LLM V2 & 35.7M & 689.5\% & 32.06\% & 10.96\% & 136.1\% & 2152\% & 163.9\% & 265.6\% & 16.69\% \\
 & Dobi-SVD & 37.0M & 2607\% & 113.3\% & 34.70\% & 209.6\% & 1795\% & 285.9\% & 278.3\% & \textbf{13.58\%} \\
 & SAES-SVD & 35.2M & 156.3\% & 18.10\% & 8.68\% & 68.40\% & 609.3\% & 74.10\% & 115.5\% & 20.65\% \\
\hdashline
 & \textbf{Ours} & 35.3M & 24.98\% & 4.23\% & \textbf{0.85\%} & 11.32\% & 127.4\% & \textbf{26.40\%} & 27.05\% & 26.91\% \\
 \rowcolor{gray!15} & \textbf{Ours*} & 35.3M & \textbf{18.92\%} & \textbf{2.90\%} & 1.30\% & \textbf{7.26\%} & \textbf{81.80\%} & 29.87\% & \textbf{26.11\%} & 27.55\% \\
\midrule
\multirow{7}{*}{0.2} & ASVD & 16.5M & 15583\% & 535.6\% & 12.51\% & 484.5\% & 48919\% & 1392\% & 3016\% & 94.46\% \\
 & FWSVD & 17.7M & 11265\% & 295.8\% & 29.67\% & 462.3\% & 43908\% & 1232\% & 2769\% & 66.10\% \\
 & SVD-LLM V2 & 18.3M & 2245\% & 67.19\% & 40.38\% & 290.2\% & 16822\% & 399.9\% & 1230\% & 34.89\% \\
 & Dobi-SVD & 20.2M & 3705\% & 118.8\% & 46.52\% & 262.2\% & 4929\% & 358.1\% & 630.2\% & \textbf{7.02\%} \\
 & SAES-SVD & 17.7M & 651.3\% & 37.96\% & 16.22\% & 165.2\% & 2952\% & 169.7\% & 432.3\% & 12.12\% \\
\hdashline
 & \textbf{Ours} & 17.8M & 291.1\% & 23.19\% & \textbf{7.87\%} & 76.27\% & 1051\% & \textbf{99.28\%} & 291.8\% & 16.10\% \\
\rowcolor{gray!15} & \textbf{Ours*} & 17.8M & \textbf{239.7\%} & \textbf{20.17\%} & 8.07\% & \textbf{66.31\%} & \textbf{927.2\%} & 116.8\% & \textbf{248.3\%} & 17.75\% \\
\bottomrule
\end{tabular}
}
\end{table*}

\subsection{Comparisons on Poseidon}
We consider three representative scenarios: NS-PwC, CE-RM, and Wave-Gauss, with results reported in Table \ref{tab:poseidon}. These tasks cover distinct physical regimes, including sharp-interface incompressible flow, shock-driven turbulent mixing, and smooth wave propagation. The Method column lists the uncompressed model, baseline methods, and our method. Since different methods may not exactly achieve the target retention ratio, the actual compressed model sizes (e.g. the values in Size column) can vary.

Across all three scenarios, our method consistently outperforms all baselines across every compression ratio.
At a target retention ratio of 0.6, the compressed models remain nearly indistinguishable from the original models across all three tasks. Even at a target retention ratio of 0.4, the degradation of our method stays below $0.5\%$ on nearly all reported metrics, demonstrating strong robustness under aggressive compression. At the most extreme setting ($20\%$ remaining parameters, ratio 0.2), performance remains strong on NS-PwC and CE-RM, while Wave-Gauss shows larger increases in Sobolev Loss, although still substantially outperforming all baselines. This trend is consistent with the underlying physics. NS-PwC and CE-RM are primarily governed by localized structures such as vortices, transport fronts, and shocks, where errors tend to remain spatially concentrated. By contrast, Wave-Gauss is dominated by globally coherent propagating modes, making it more sensitive to small compression-induced perturbations that accumulate as phase shifts and wave-speed errors.

\subsection{Comparisons on VICON}
Unlike Poseidon, which is fine-tuned separately on each dataset prior to compression, VICON is a unified model trained across multiple physical domains. To account for this multi-domain setting, we evaluate two variants of our method: Ours and Ours*. Both use fused multi-scenario calibration, while Ours* further reweights dataset contributions using energy balance to mitigate cross-domain scale disparities during compression. Further details are provided in the Appendix \ref{app:trace_balance}. 

Evaluation results are summarized in Table \ref{tab:vicon}. Overall, our method consistently achieves the best or near-best trade-off across datasets and compression levels. Under aggressive compression (ratio 0.2), all methods suffer a sharp increase in Base Loss, indicating that the remaining model capacity is insufficient for reliable deployment. Although some baselines achieve lower Sobolev errors in specific cases (e.g., Euler-2D), they come with substantially higher Base Loss, limiting their practical value. In contrast, our method maintains low Base Loss while remaining competitive on Sobolev metrics, resulting in stronger overall solution quality. While Ours outperforms Ours* on certain metrics at specific compression ratios, Ours* delivers better overall performance across settings.

\subsection{Qualitative Comparison}
\begin{figure}[tb]
    \centering
    \begin{subfigure}{\textwidth}
        \includegraphics[width=\textwidth]{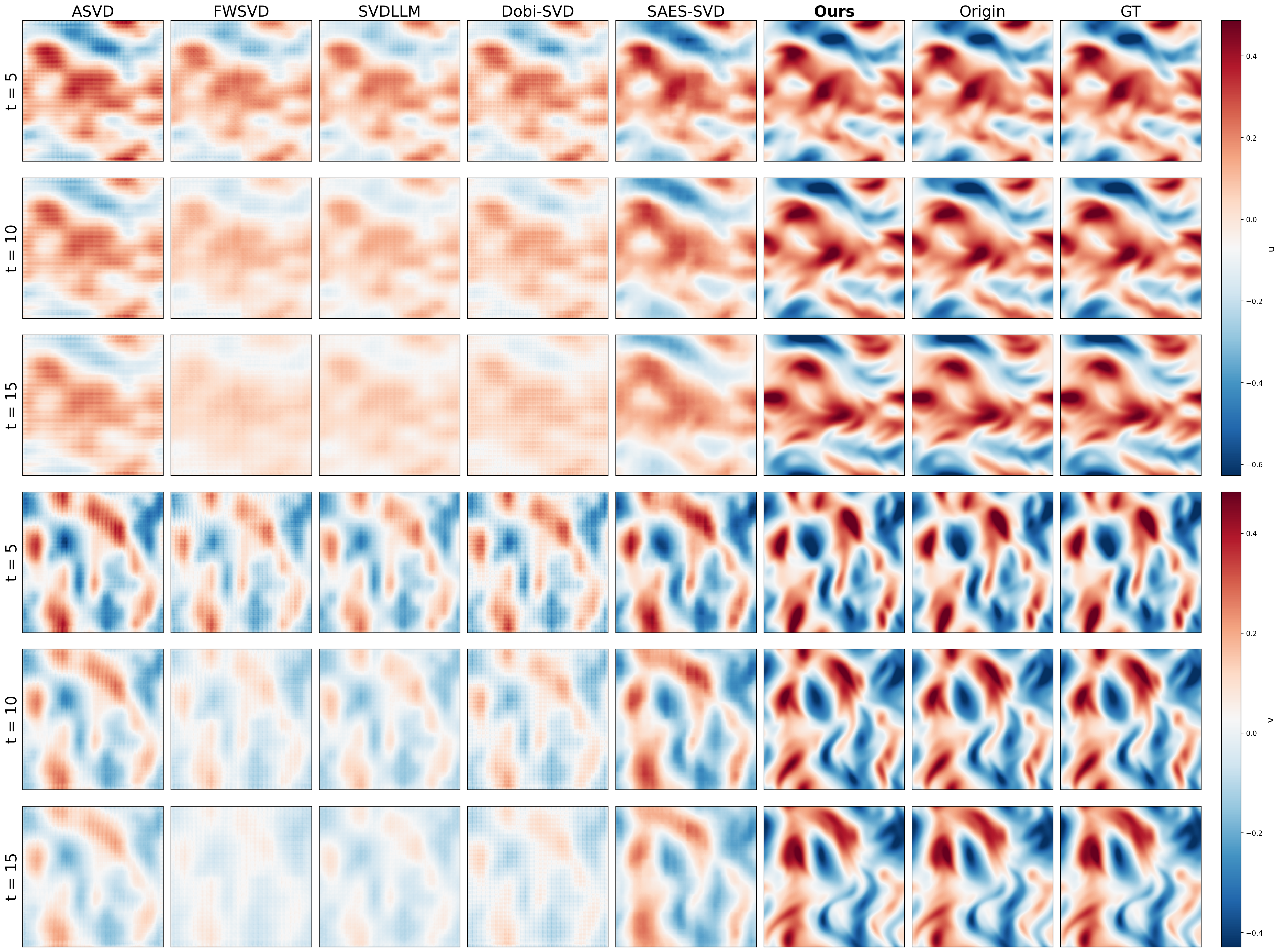}
        \label{fig: quali_NS_PwC}
    \end{subfigure}
    \caption{Predicted velocity fields ($u, v$) on the NS-PwC dataset at
    target retention ratio 0.2 over three rollout horizons ($t=5, 10, 15$).}
    \label{fig:quali_ns_pwc}
\end{figure}
\paragraph{NS-PwC}
\cref{fig:quali_ns_pwc} compares predicted velocity fields ($u, v$) on the
NS-PwC (Navier--Stokes, Piecewise Constants) dataset.  This dataset features
turbulent vortex dynamics driven by piecewise-constant initial conditions,
producing sharp shear layers and concentrated vortex cores that evolve over
time.

Most baselines, FWSVD, SVD-LLM V2, and Dobi-SVD, exhibit severe
over-smoothing which progressively erase fine-scale vortical structures; by
$t{=}15$, their predictions become largely featureless, with Dobi-SVD
additionally displaying noticeable checkerboard artifacts.  ASVD and SAES-SVD preserve
more structure than these baselines but still loses sharp gradients and
concentrated vortex cores.  In contrast, our method remains visually consistent
with the original model across all horizons, accurately preserving vortex-pair
advection, shear-layer sharpness, and the global energy distribution.

These observations align with the quantitative results in \Cref{tab:poseidon}:
at a target retention ratio of 0.2, baselines incur Sobolev increases of 2429--10090\%,
indicating massive gradient degradation, while our method achieves below 3.0\%
increase in both Base Loss and Sobolev norm.

\begin{figure}[tb]
    \centering
    \begin{subfigure}{\textwidth}
        \includegraphics[width=\textwidth]{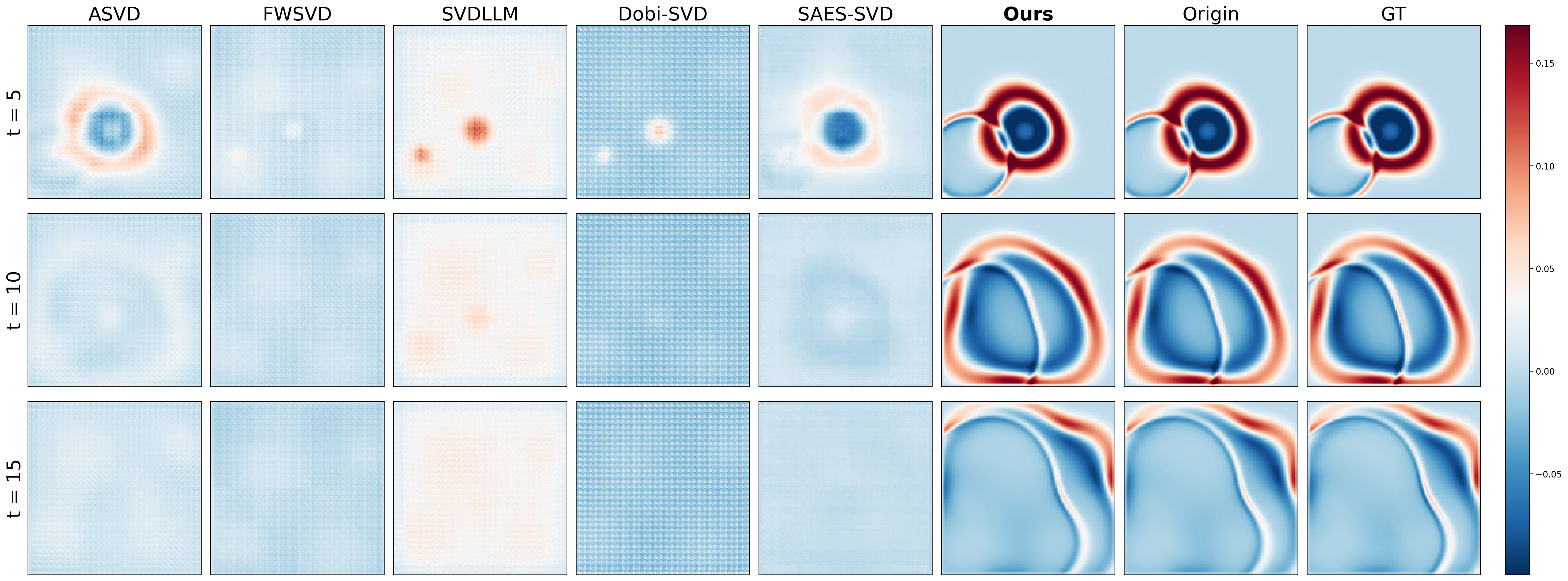}
        \label{fig: quali_Wave_Gauss}
    \end{subfigure}
    \caption{Predicted wave field $u$ on the Wave-Gauss dataset at target retention
    ratio 0.2 over three rollout horizons ($t=5, 10, 15$).}
    \label{fig:quali_wave_gauss}
    \vspace{-1em}
\end{figure}

\paragraph{Wave-Gauss}
We show one visual comparison in \Cref{fig:quali_wave_gauss} on the Wave-Gauss dataset. This dataset contains wave propagation from Gaussian initial conditions, producing expanding concentric wavefronts that develop complex
interference patterns over time. FWSVD, SVD-LLM V2, and Dobi-SVD fail to resolve the expanding circular wavefronts at $t{=}5$ already and degrade to essentially constant or artifact-dominated outputs by $t{=}15$.  ASVD and SAES-SVD capture the gross wavefront geometry but loses the fine-scale interference patterns.  Our method faithfully reproduces both the sharp wavefront edges and the intricate interference structures across all rollout steps, closely tracking the original model. These qualitative results also align with the results in \Cref{tab:poseidon}. At a target retention ratio of 0.2, baselines exhibit
Base Loss increases of 366.9--1122\% and Sobolev increases ranging from 5,873\% to 20,124\%, reflecting a near-total loss of both amplitude and gradient
information.  Our method maintains a Base Loss increase of only 0.70\% and a
Sobolev increase of 152\%.

\section{Conclusion, Limitation, and Future Work}
\label{sec:conclusion}
We presented the first physics-aware SVD compression framework, SAFE-SVD, for physics foundation models. As PFMs have started a fundamental shift in AI for Science, SAFE-SVD is timely and proactive to anticipate and address the imminent energy bottleneck caused by ever-growing PFMs. SAFE-SVD considers both predictive accuracy and physical fidelity during compression, which are built in the modeling of layer sensitivity, cross-layer error propagation, and adaptive rank allocation. Experiments on three major PFMs show massive and consistent improvements over existing baselines across diverse PDE tasks and physical metrics. For example, in Poseidon on CE-RM and NS-PwC, when we specify 2.5\% on CE-RM and 3\% on NS-PwC as the post-compression error-change threshold, SAFE-SVD only needs 20\% of the original model size, while SAES-SVD requires 60\% of the original model size on CE-RM, and no baseline meets threshold on NS-PwC.

One limitation is SAFE-SVD is specific to PDE families. This is because we need the loss terms in the Sobolev space where the orders of derivatives are decided by the specific PDE family in the calibration dataset. However, we argue that a calibration dataset is always available in practice as the target application domain is usually given. Moreover, for applications requiring high precision, compression-induced errors may limit the applicability of our method. However, this is a common problems for all compression methods. Also, some applications do not require high-precision predictions and can benefit from model compression; For example, building ventilation design relies on aerodynamic simulations, where engineering decisions are often based on relative ventilation performance rather than exact flow predictions \cite{hwang2022large}. Consequently, moderate approximation errors can often be tolerated, allowing compressed physics foundation models to serve as efficient surrogate models that substantially reduce computational cost.

An important direction for future work is to incorporate stronger physics priors, such as conservation laws, symmetries, and geometry-aware constraints. Another promising direction is extending the framework to larger-scale and multimodal scientific foundation models.

\section{Acknowledgment}
This work was supported in part by the UK Research and Innovation AIRR Innovator Award (0261-5654-9320-1), and the Natural Science Foundation of Hunan Province (2026JJ60508).

\bibliographystyle{plainnat}
\bibliography{references}

\newpage
\appendix
\etocsettocstyle{\section*{Appendix}}{} 
\localtableofcontents

\section{Detailed Derivation of Physics-Aware Compression}
\label{app:detailed_derivation}
This appendix provides the preliminaries, detailed derivations and rank selection reasoning omitted in the main text. We first introduce the necessary preliminaries, then show how the proposed loss-aware compression objective is derived and demonstrate how the resulting optimization can be reduced to a standard low-rank approximation problem. Finally, we reason why our rank selection is efficient. Throughout, we focus on a single layer and omit the layer index for clarity.

\subsection{Preliminaries}
\paragraph{Low-Rank Compression via SVD}
\label{app:loraviasvd}
Given a matrix $\mathbf{W} \in \mathbb{R}^{m \times n}$, its Singular Value Decomposition is $\mathbf{W} = \mathbf{U}\mathbf{\Sigma}\mathbf{V}^{\top}$, 
where $\mathbf{U} \in \mathbb{R}^{m \times m}$ and $\mathbf{V} \in \mathbb{R}^{n \times n}$ are orthogonal matrices, and $\mathbf{\Sigma} \in \mathbb{R}^{m \times n}$ is diagonal with singular values ordered as $\sigma_1 \ge \sigma_2 \ge \dots \ge \sigma_r > 0$. A rank-$k$ approximation is obtained by truncating the spectrum,
yielding $\mathbf{W} \approx \mathbf{W}_k
= \mathbf{U}_k\mathbf{\Sigma}_k\mathbf{V}_k^\top$,
where $\mathbf{U}_k \in \mathbb{R}^{m\times k}$ and
$\mathbf{V}_k \in \mathbb{R}^{n\times k}$ contain the top $k$
left and right singular vectors, respectively, and
$\mathbf{\Sigma}_k = \operatorname{diag}(\sigma_1,\ldots,\sigma_k)
\in \mathbb{R}^{k\times k}$ contains the corresponding singular values.
By the Eckart--Young--Mirsky theorem, $\mathbf{W}_k$ is an optimal
rank-$k$ approximation under the Frobenius norm.
Furthermore, $\mathbf{W}_k$ can be rewritten as
$(\mathbf{U}_k\mathbf{\Sigma}_k^{1/2})
(\mathbf{\Sigma}_k^{1/2}\mathbf{V}_k^\top)$.
These two factors have dimensions $m\times k$ and $k\times n$,
respectively, requiring $k(m+n)$ parameters instead of $mn$.

\paragraph{Fisher Information}
\label{app:fisher_matrix_definition}
Following empirical approximation, the (empirical) FIM used in our paper is computed as
\begin{equation}
\label{eq:FIM_def}
    \mathbf{F}_{\mathbf{Z}}
    =
    \frac{1}{N_{\mathrm{cal}}}
    \sum_{i=1}^{N_{\mathrm{cal}}}
    \mathbf{g}_i \mathbf{g}_i^\top,
    \qquad
    \mathbf{g}_i
    =
    \nabla_{\mathbf{Z_i}} \mathcal{L}(\mathbf{Z_i}).
\end{equation}
where $N_{cal}$ is the number of calibration samples.

\subsection{The loss change formulation}
\label{app:theory_alignment}

The choice between an $l_1$ and an $l_2$ Sobolev loss does not
change the derivation of our final objective. Following BRECQ
\cite{li2021brecq}, the difference between the compressed-model loss
and the original-model loss can be approximated using a second-order
Taylor expansion:
\begin{equation}
\Delta \mathcal{L}
=
\mathcal{L}(\mathbf{W}+\Delta\mathbf{W})
-
\mathcal{L}(\mathbf{W})
\approx
\mathbf{g}_{W}^{\top}\Delta\mathbf{W}
+
\frac{1}{2}
\Delta\mathbf{W}^{\top}
\mathbf{H}_{W}
\Delta\mathbf{W},
\end{equation}
where $\mathcal{L}$ does not prescribe a specific form.
In our setting, $\mathcal{L}$ includes the Sobolev loss, which may
use either an $l_1$ or an $l_2$ criterion.

Following the BRECQ assumption that the pretrained model is
sufficiently well trained, the expected first-order term is
negligible, leading to
\begin{equation}
\min_{\Delta\mathbf{W}}
\mathbb{E}[\Delta\mathcal{L}]
\approx
\min_{\Delta\mathbf{W}}
\frac{1}{2}
\mathbb{E}
\left[
\Delta\mathbf{W}^{\top}
\mathbf{H}_{W}
\Delta\mathbf{W}
\right].
\end{equation}

Under local linearization,
\begin{equation}
\Delta\mathbf{Z}
\approx
\mathbf{J}_{Z}\Delta\mathbf{W},
\qquad
\mathbf{H}_{W}
\approx
\mathbf{J}_{Z}^{\top}
\mathbf{H}_{Z}
\mathbf{J}_{Z},
\end{equation}
which gives
\begin{equation}
\Delta\mathbf{W}^{\top}
\mathbf{H}_{W}
\Delta\mathbf{W}
\approx
\Delta\mathbf{Z}^{\top}
\mathbf{H}_{Z}
\Delta\mathbf{Z}.
\end{equation}

Approximating $\mathbf{H}_{Z}$ using the Fisher Information Matrix
then yields the quadratic activation-space objective used in
Equation~\ref{eq:objective_ori}.
This approximation interprets the objective as a composite
log-likelihood, where the $l_1$ and $l_2$ terms correspond to
Laplace and Gaussian distributions, respectively.

\subsection{From Base Objective to Trace Formulation}
\label{app:Frobenius_to_trace}
Starting from the compression objective in Equation \ref{eq:full_objective} in the main paper, we optimize
\begin{equation}
\label{eq:appendix_full_objective}
    \min_{ \mathbf{W}'} \,\, (1-\alpha) \mathbb{E} \left[ \| \mathbf{L}(\mathbf{W}\mathbf{X} - \mathbf{W}'\mathbf{X}) \|_2^2 \right] + \alpha \mathbb{E} \left[ \| \mathbf{L}(\mathbf{W}\mathbf{X} - \mathbf{W}'\mathbf{X}') \|_2^2 \right],
\end{equation}
where $\mathbf{X}$ denotes the clean input activation, $\mathbf{X}'$ is the shifted activation induced by previously compressed layers, and $\mathbf{L}$ is decomposed from $\mathbf{F_Z}$ (Equation~\ref{eq:FIM_def}), \ie $\mathbf{F_Z} = \mathbf{L}^\top \mathbf{L}$. $\mathbf{W}$ and $\mathbf{W}'$ are the original and compressed weights, respectively. Note that the expectation $\mathbb{E}[\cdot]$ is taken over the data distribution. Therefore, the activations $\mathbf{X}$ and $\mathbf{X}'$ are random variables, while $\mathbf{W}$, $\mathbf{W}'$, and $\mathbf{L}$ which is estimated over the entire calibration set, are deterministic constants with respect to this expectation. The first term captures intra-layer reconstruction error under clean inputs, while the second term models propagated error caused by preceding layer compression.

For a single-sample vector $\mathbf{a}$, we use
\begin{equation}
    \|\mathbf{a}\|_2^2
    =
    \operatorname{Tr}(\mathbf{a}\mathbf{a}^{\top}).
\end{equation}

For the first term, using the linearity of the trace operator and the deterministic nature of $\mathbf{L}$, $\mathbf{W}$, and $\mathbf{W}'$, we can move the expectation inside:
\begin{equation}
\begin{aligned}
    \mathbb{E} [\| \mathbf{L}(\mathbf{W}\mathbf{X} & -  \mathbf{W}'\mathbf{X}) \|_2^2 ] = \mathbb{E} \left[ \operatorname{Tr} \left( \mathbf{L}(\mathbf{W}\mathbf{X} - \mathbf{W}'\mathbf{X})(\mathbf{W}\mathbf{X} - \mathbf{W}'\mathbf{X})^\top \mathbf{L}^\top \right) \right] \\
    & = \mathbb{E} \left[ \operatorname{Tr} \left( \mathbf{L}(\mathbf{W} - \mathbf{W}')\mathbf{X}\mathbf{X}^\top(\mathbf{W} - \mathbf{W}')^\top\mathbf{L}^\top \right) \right] \\
    & = \operatorname{Tr} \left( \mathbf{L}(\mathbf{W} - \mathbf{W}')\mathbb{E}[\mathbf{X}\mathbf{X}^\top](\mathbf{W} - \mathbf{W}')^\top\mathbf{L}^\top \right).
\end{aligned}
\end{equation}

Let $\Sigma_{\mathbf{X}\mathbf{X}}=\mathbb{E}\left[ \mathbf{X}\mathbf{X}^\top \right]$, then expanding the quadratic form gives
\begin{equation}
\label{eq:first_term_expanded}
\begin{aligned}
   \mathbb{E} [ \| \mathbf{L}(\mathbf{W}\mathbf{X} & - \mathbf{W}'\mathbf{X}) \|_2^2 ] = \operatorname{Tr}(\mathbf{L}\mathbf{W}\boldsymbol{\Sigma}_{\mathbf{X}\mathbf{X}}\mathbf{W}^\top\mathbf{L}^\top) - \operatorname{Tr}(\mathbf{L}\mathbf{W}\boldsymbol{\Sigma}_{\mathbf{X}\mathbf{X}}\mathbf{W}'^\top\mathbf{L}^\top) \\
    &- \operatorname{Tr}(\mathbf{L}\mathbf{W}'\boldsymbol{\Sigma}_{\mathbf{X}\mathbf{X}}\mathbf{W}^\top\mathbf{L}^\top) + \operatorname{Tr}(\mathbf{L}\mathbf{W}'\boldsymbol{\Sigma}_{\mathbf{X}\mathbf{X}}\mathbf{W}'^\top\mathbf{L}^\top),
\end{aligned}
\end{equation}
Since $\operatorname{Tr}(\mathbf{L}\mathbf{W}\boldsymbol{\Sigma}_{\mathbf{X}\mathbf{X}}\mathbf{W}^\top\mathbf{L}^\top)$ is independent to $\mathbf{W}'$ and using transpose invariance of the trace $\operatorname{Tr}(\mathbf{A}) = \operatorname{Tr}(\mathbf{A}^\top)$, we simplify:
\begin{equation}
\label{eq:first_term_simplified}
    \mathbb{E} \left[ \| \mathbf{L}(\mathbf{W}\mathbf{X} - \mathbf{W}'\mathbf{X}) \|_2^2 \right] = \text{const} - 2\operatorname{Tr}(\mathbf{L}\mathbf{W}'\boldsymbol{\Sigma}_{\mathbf{X}\mathbf{X}}\mathbf{W}^\top\mathbf{L}^\top) + \operatorname{Tr}(\mathbf{L}\mathbf{W}'\boldsymbol{\Sigma}_{\mathbf{X}\mathbf{X}}\mathbf{W}'^\top\mathbf{L}^\top).
\end{equation}
Similarly, the second term can be expressed as
\begin{equation}
\label{eq:second_term}
    \mathbb{E} \left[ \| \mathbf{L}(\mathbf{W}\mathbf{X} - \mathbf{W}'\mathbf{X}') \|_2^2 \right] =  \text{const} - 2\operatorname{Tr}(\mathbf{L}\mathbf{W}'\boldsymbol{\Sigma}_{\mathbf{X}^{'}\mathbf{X}}\mathbf{W}^\top\mathbf{L}^\top) + \operatorname{Tr}(\mathbf{L}\mathbf{W}'\boldsymbol{\Sigma}_{\mathbf{X}^{'}\mathbf{X}^{'}}\mathbf{W}'^\top\mathbf{L}^\top),
\end{equation}
where 
\begin{equation}
\nonumber
\boldsymbol{\Sigma}_{\mathbf{X}^{'}\mathbf{X}} = \mathbb{E}[\mathbf{X}{'}\mathbf{X}^\top], \quad \boldsymbol{\Sigma}_{\mathbf{X}^{'}\mathbf{X}^{'}} = \mathbb{E}[\mathbf{X}'\mathbf{X}'^\top].
\end{equation}
Substituting Equation \ref{eq:first_term_simplified} and Equation \ref{eq:second_term} into 
Equation \ref{eq:appendix_full_objective} and collecting all terms that depend on $\mathbf{W}'$, we obtain
\begin{align}
\label{eq:appendix_full_objective_trace_form}
    \min_{ \mathbf{W}'} \,\, (1-\alpha) \mathbb{E} \left[ \| \mathbf{L}(\mathbf{W}\mathbf{X} - \mathbf{W}'\mathbf{X}) \|_2^2 \right] + \alpha \mathbb{E} \left[ \| \mathbf{L}(\mathbf{W}\mathbf{X} - \mathbf{W}'\mathbf{X}') \|_2^2 \right] \\
    =\min_{ \mathbf{W}'} \,\,  -2\operatorname{Tr}\left(\mathbf{L}\mathbf{W}'\boldsymbol{\Sigma}_{\text{cov}}\mathbf{W}^\top\mathbf{L}^\top\right) + \operatorname{Tr}\left(\mathbf{L}\mathbf{W}'\mathbf{\Sigma}_{\text{c\_cov}}\mathbf{W}'^\top\mathbf{L}^\top\right) + \text{const},
\end{align}
with covariance matrices
\begin{equation}
    \mathbf{\Sigma}_{\text{cov}}= (1-\alpha)\boldsymbol{\Sigma}_{\mathbf{X}\mathbf{X}} + \alpha\boldsymbol{\Sigma}_{\mathbf{X}^{'}\mathbf{X}^{}},\quad \mathbf{\Sigma}_{\text{c\_cov}}= (1-\alpha)\boldsymbol{\Sigma}_{\mathbf{X}\mathbf{X}} + \alpha\boldsymbol{\Sigma}_{\mathbf{X}^{'}\mathbf{X}^{'}}.
\end{equation}
This is the trace-equivalent form used in Equation \ref{eq:full_objective_trace_form} of the main paper. It integrates intra-layer reconstruction perturbation, and propagated compression error into a single unified objective.

\subsection{From Trace Formulation to Frobenius-norm}
\label{app:trace_to_Frobenius}
To obtain a tractable solver, we convert Equation \ref{eq:appendix_full_objective_trace_form} into a standard low-rank regression problem. 
In practice, $\mathbf{\Sigma}_{\text{c\_cov}}$ is positive definite Appendix \ref{app:pd}, allowing a Cholesky decomposition $\boldsymbol{\Sigma}_{\text{c\_cov}} = \mathbf{R}\mathbf{R}^\top$. Substituting this factorization into the second term in Equation \ref{eq:appendix_full_objective_trace_form} yields:
\begin{equation}
    \operatorname{Tr}(\mathbf{L}\mathbf{W}'\mathbf{R}\mathbf{R}^\top\mathbf{W}'^\top\mathbf{L}^\top) = 
    \operatorname{Tr}(\mathbf{L}\mathbf{W}'\mathbf{R}(\mathbf{L}\mathbf{W}'\mathbf{R})^\top).
\end{equation}

For the first term in Equation \ref{eq:appendix_full_objective_trace_form}, we rearrange it
\begin{equation}
\begin{aligned}
    -2\operatorname{Tr}(\mathbf{L}\mathbf{W}'&{\Sigma}_{\text{cov}}\mathbf{W}^\top\mathbf{L}^\top)  = -2\operatorname{Tr}(\mathbf{L}\mathbf{W}'\mathbf{R}\mathbf{R}^{-1}{\Sigma}_{\text{cov}}\mathbf{W}^\top\mathbf{L}^\top) \\
    &= -2\operatorname{Tr}\big( (\mathbf{R}^{-1}{\Sigma}_{\text{cov}}\mathbf{W}^\top\mathbf{L}^\top) (\mathbf{L}\mathbf{W}'\mathbf{R}) \big) \\
    &= -2\operatorname{Tr}\big( (\mathbf{L}\mathbf{W}{\Sigma}^\top_{\text{cov}}\mathbf{R^{-\top}})^\top (\mathbf{L}\mathbf{W}'\mathbf{R}) \big).
\end{aligned}
\end{equation}

For any matrices $\mathbf{A}$ and $\mathbf{B}$ of compatible dimensions, the following holds:
\begin{equation}
||\mathbf{A} - \mathbf{B}||_F^2 = \operatorname{Tr}(\mathbf{A}\mathbf{A}^\top) - 2\operatorname{Tr}(\mathbf{B}^\top\mathbf{A}) + ||\mathbf{B}||_F^2.
\end{equation}
We define 
\begin{equation}
    \nonumber
    \mathbf{A} = {\mathbf{L}\mathbf{W}'\mathbf{R}}, \quad \mathbf{B} = {\mathbf{L}\mathbf{W}\boldsymbol{\Sigma}_{\text{cov}}^\top \mathbf{R}^{-\top}}.
\end{equation}
Since $\|\mathbf{B}\|_F^2$ does not depend on $\mathbf{W}'$, it can be absorbed into the $\mathrm{const}$ term, allowing us to rewrite the objective as:

\begin{equation}
    \min_{ \mathbf{W}'} \,\, \left\| \mathbf{L}\mathbf{W}'\mathbf{R} - \mathbf{L}\mathbf{W}\mathbf{\Sigma}^\top_{\text{cov}}\mathbf{R}^{-\top} \right\|_F^2.
\end{equation}

This objective is exactly the same as Equation \ref{eq:Frobenius_norm_regression} in the main paper.

\subsection{Derivation of Loss Degradation for Rank Selection}
\label{appendix: loss degradation}
As derived from Equation \ref{eq:objective}, the impact of the perturbation $\Delta \mathbf{Z}$ on the final loss can be approximated by the objective:
\begin{equation}
    \mathbb{E}[\Delta\mathcal{L}]
    \approx
    {1 \over 2}\mathbb{E}
    \left[
    \| \mathbf{L}\Delta\mathbf{Z} \|_2^2
    \right].
\end{equation}
If we strictly consider the clean input condition (i.e., isolating the current layer by assuming no upstream errors, $\mathbf{X}' = \mathbf{X}$), the perturbation simplifies to $\Delta \mathbf{Z} = (\mathbf{W'} - \mathbf{W})\mathbf{X}$. Let $\mathbf{A} = \mathbf{L}(\mathbf{W'} - \mathbf{W})$ for brevity. We expand the expected squared norm:
\begin{equation}
    \mathbb{E} \left[ \| \mathbf{A}\mathbf{X} \|_2^2 \right]
    = \mathbb{E} \left[ \mathrm{tr}\!\left( \mathbf{A}\mathbf{X}\mathbf{X}^\top \mathbf{A}^\top \right) \right]
    = \mathrm{tr}\!\left( \mathbf{A}\, \mathbb{E}[\mathbf{X}\mathbf{X}^\top]\, \mathbf{A}^\top \right)
    = \mathrm{tr}\!\left( \mathbf{A}\, \boldsymbol{\Sigma}_{\mathbf{XX}}\, \mathbf{A}^\top \right).
\end{equation}
Recalling that the input covariance is decomposed as $\boldsymbol{\Sigma}_{\mathbf{XX}} = \mathbf{R}^c\mathbf{R}^{c\top}$, we substitute and apply the identity $\mathrm{tr}(\mathbf{B}\mathbf{B}^\top) = \|\mathbf{B}\|_F^2$:
\begin{equation}
    \mathrm{tr}\!\left( \mathbf{A}\, \mathbf{R}^c\mathbf{R}^{c\top}\, \mathbf{A}^\top \right)
    = \mathrm{tr}\!\left( (\mathbf{A}\mathbf{R}^c)(\mathbf{A}\mathbf{R}^c)^\top \right)
    = \| \mathbf{A}\mathbf{R}^c \|_F^2
    = \| \mathbf{L}\mathbf{W'}\mathbf{R}^c - \mathbf{L}\mathbf{W}\mathbf{R}^c \|_F^2.
\end{equation}
Thus under clean input condition, the loss change can be approximated:
\begin{equation}
    \mathbb{E}\left[ \Delta \mathcal{L}\right] \approx
    {1 \over 2}\mathbb{E} \left[ \| \mathbf{L} (\mathbf{W'} - \mathbf{W})\mathbf{X} \|_2^2 \right] = {1 \over 2}\| \mathbf{L}\mathbf{W'}\mathbf{R}^c - \mathbf{L}\mathbf{W}\mathbf{R}^c \|_F^2.
    \label{eq:whitened_frob}
\end{equation}
Let $\mathbf{M}^* = \mathbf{L}\mathbf{W}\mathbf{R}^c$ and $\mathbf{M} = \mathbf{L}\mathbf{W}'\mathbf{R}^c$. The compression objective becomes a standard low-rank approximation problem:
\begin{equation}
    \min_{\mathbf{M}} {1 \over 2}\| \mathbf{M} - \mathbf{M}^* \|_F^2 \quad \text{subject to} \quad \mathrm{rank}(\mathbf{M}) = k.
\end{equation}
By the Eckart--Young--Mirsky theorem, the optimal solution is the rank-$k$ truncated SVD of $\mathbf{M}^*$.
Writing $\mathbf{M}^* = \mathbf{U}\boldsymbol{\Sigma}\mathbf{V}^\top$, if we decide to retain the top-$k$ ranks, the residual error equals the sum of the squared singular values beyond rank $k$.
Thus, the approximated loss degradation is:
\begin{equation}
\mathbb{E}[\Delta\mathcal{L}]
\approx
\frac{1}{2}
\mathbb{E}\!\left[
\|L(W'-W)X\|_2^2
\right]
=
\frac{1}{2}
\sum_{i=k+1}^{r}\sigma_i^2.
\end{equation}
where $r$ denotes the full rank of the transformed matrix, \ie $r=rank(\mathbf{M^*})=rank(\mathbf{W})$ and $\sigma_{i}$ is its $i$-th singular value.
Consequently, under the clean-input quadratic approximation,
dropping the $i$-th singular component contributes $\sigma_i^2/2$ to the estimated loss degradation. This provides a unified, globally comparable effect across all heterogeneous layers, \ie all linear layers can be compared under the same evaluation.
\section{Additional Experimental Results}
\label{app:additional_exp_res}
\subsection{Quantitative Results}
We provide fully detailed experimental results. For all comparisons, we tested target retention ratios 1.0, 0.8, 0.6, 0.4, and 0.2. Some results are presented in the main paper, but we still list them for the purpose of easy comparisons.
\subsubsection{Comparisons on Poseidon}
The results on the additional datasets, as shown in Table \ref{tab:poseidon_appendix}, further solidify the robustness and generalizability of our proposed method across diverse physical domains. Across all three scenarios---NS-BB, CE-RPUI, and Wave-Layer---our method consistently maintains a significant lead over all baselines. While traditional SVD-based approaches typically experience catastrophic performance collapse (with errors frequently exceeding 100\% or even 1000\%) as soon as the target retention ratio drops below 0.8, our method keeps the error near-zero at moderate compression and preserves physical validity even at the extreme 0.2 retention ratio.

\paragraph{NS-BB} 
In the Navier-Stokes Brownian Bridge(NS-BB) scenario, which involves complex multi-scale eddy structures, our method demonstrates exceptional adherence to physical constraints. Even at the most aggressive target retention ratio of 0.2, the divergence-free (Div-Free) and vorticity errors remain remarkably low at \textbf{0.29\%} and \textbf{0.77\%}, respectively. This indicates that our approach effectively captures the underlying conservation laws and local fluid dynamics, preventing the accumulation of unphysical artifacts that plague baseline methods.

\paragraph{CE-RPUI} 
The Compressible Euler Riemann-Kelvin-Helmholtz (CE-RPUI) task presents a significant challenge due to its shock-driven turbulent mixing and sensitivity to interface discontinuities. Our method exhibits high robustness in this regime; at a 0.4 ratio, the degradation remains negligible, and even at the 0.2 limit, the Base Loss is contained at \textbf{4.50\%}. In contrast, baseline methods like ASVD and SAES-SVD fail to maintain the integrity of the shock fronts, resulting in errors that are orders of magnitude higher.

\paragraph{Wave-Layer} 
For the Wave-Layer dataset, which simulates wave propagation through stratified media, the results reinforce our observations regarding wave-dominated physics. As with the Wave-Gaussian task in the main text, the Sobolev Loss shows greater sensitivity at extreme compression (reaching \textbf{41.07\%} at ratio 0.2), which is characteristic of the phase-shift accumulation inherent in globally coherent propagating modes. However, compared to the closest baseline (SAES-SVD) which reaches a staggering \textbf{9305\%} error, our method provides a vastly more stable and physically meaningful approximation of the wave operator.
\begin{table*}[t]
\centering
\caption{Rest Experimental Results on Poseidon.}
\label{tab:poseidon_appendix}
\resizebox{\textwidth}{!}{
\begin{tabular}{c c c r r r r r r r r}
\toprule
\multirow{2}{*}{\textbf{$\eta$}} &
\multirow{2}{*}{\textbf{Method}} &
\multirow{2}{*}{\textbf{Size}} &
\multicolumn{4}{c}{\textbf{NS-BB}} &
\multicolumn{2}{c}{\textbf{CE-RPUI}} &
\multicolumn{2}{c}{\textbf{Wave-Layer}} \\
\cmidrule(lr){4-7}
\cmidrule(lr){8-9}
\cmidrule(lr){10-11}

& & &
\textbf{Base Loss $\downarrow$} &
\textbf{Sobolev $\downarrow$} &
\textbf{Div-Free $\downarrow$} &
\textbf{Vorticity $\downarrow$} &
\textbf{Base Loss $\downarrow$} &
\textbf{Sobolev $\downarrow$} &
\textbf{Base Loss $\downarrow$} &
\textbf{Sobolev $\downarrow$} \\
\midrule
1 & Origin & 628.6M & 0.00\% & 0.00\% & 0.00\% & 0.00\% & 0.00\% & 0.00\% & 0.00\% & 0.00\% \\
\midrule
\multirow{6}{*}{0.8} & ASVD & 502.8M & 1239\% & 961.8\% & 383.5\% & 1260\% & 66.56\% & 16.26\% & 53.87\% & 1946\% \\
 & FWSVD & 502.7M & 5929\% & 3264\% & 1440\% & 4073\% & 398.4\% & 70.72\% & 879.8\% & 9019\% \\
 & SVD-LLM V2 & 522.9M & 5793\% & 3495\% & 1742\% & 4061\% & 416.4\% & 53.06\% & 877.0\% & 7754\% \\
 & Dobi-SVD & 503.7M & 6974\% & 3400\% & 1322\% & 4242\% & 480.8\% & 97.35\% & 1004\% & 11207\% \\
 & SAES-SVD & 502.7M & 2254\% & 2235\% & 1297\% & 1801\% & 185.2\% & 45.63\% & 336.4\% & 5012\% \\
\hdashline
 \rowcolor{gray!15} & \textbf{Ours} & 503.0M & \textbf{0.00\%} & \textbf{0.00\%} & \textbf{0.00\%} & \textbf{0.00\%} & \textbf{0.00\%} & \textbf{0.00\%} & \textbf{0.00\%} & \textbf{0.00\%} \\
\midrule
\multirow{6}{*}{0.6} & ASVD & 377.0M & 3763\% & 3695\% & 1785\% & 3176\% & 324.8\% & 58.16\% & 386.9\% & 11953\% \\
 & FWSVD & 377.1M & 6121\% & 4058\% & 1775\% & 4148\% & 413.2\% & 118.8\% & 903.7\% & 10956\% \\
 & SVD-LLM V2 & 424.8M & 6904\% & 3677\% & 1797\% & 4353\% & 441.7\% & 121.3\% & 892.6\% & 10578\% \\
 & Dobi-SVD & 378.1M & 6785\% & 4128\% & 1718\% & 4258\% & 527.1\% & 116.9\% & 1025\% & 9618\% \\
 & SAES-SVD & 377.1M & 2480\% & 2372\% & 1372\% & 1992\% & 198.9\% & 56.34\% & 324.5\% & 6126\% \\
\hdashline
\rowcolor{gray!15} & \textbf{Ours} & 377.5M & \textbf{0.00\%} & \textbf{0.01\%} & \textbf{0.01\%} & \textbf{0.01\%} & \textbf{0.04\%} & \textbf{0.02\%} & \textbf{0.01\%} & \textbf{0.01\%} \\
\midrule
\multirow{6}{*}{0.4} & ASVD & 251.4M & 4691\% & 6606\% & 2967\% & 3998\% & 441.4\% & 207.9\% & 627.5\% & 22672\% \\
 & FWSVD & 251.5M & 7133\% & 7037\% & 2926\% & 4383\% & 446.1\% & 323.2\% & 940.5\% & 14994\% \\
 & SVD-LLM V2 & 321.6M & 7242\% & 5031\% & 2371\% & 4455\% & 491.2\% & 280.1\% & 1021\% & 28063\% \\
 & Dobi-SVD & 252.6M & 9064\% & 7041\% & 3003\% & 4564\% & 511.0\% & 186.6\% & 1037\% & 22288\% \\
 & SAES-SVD & 251.5M & 2641\% & 2535\% & 1486\% & 2095\% & 213.4\% & 72.77\% & 324.1\% & 6886\% \\
\hdashline
\rowcolor{gray!15} & \textbf{Ours} & 251.9M & \textbf{0.10\%} & \textbf{0.05\%} & \textbf{0.01\%} & \textbf{0.08\%} & \textbf{0.15\%} & \textbf{0.14\%} & \textbf{0.07\%} & \textbf{0.31\%} \\
\midrule
\multirow{6}{*}{0.2} & ASVD & 125.8M & 6179\% & 8106\% & 3484\% & 4777\% & 534.9\% & 555.4\% & 721.2\% & 22274\% \\
 & FWSVD & 125.9M & 7644\% & 11629\% & 4266\% & 4761\% & 523.6\% & 592.4\% & 1047\% & 25799\% \\
 & SVD-LLM V2 & 202.7M & 8190\% & 6519\% & 3020\% & 4701\% & 623.8\% & 288.4\% & 1719\% & 27723\% \\
 & Dobi-SVD & 127.0M & 7138\% & 11119\% & 2838\% & 4902\% & 582.3\% & 364.8\% & 2447\% & 26380\% \\
 & SAES-SVD & 125.9M & 3383\% & 3349\% & 1914\% & 2692\% & 270.0\% & 96.17\% & 504.9\% & 9305\% \\
\hdashline
\rowcolor{gray!15} &  \textbf{Ours} & 126.3M & \textbf{1.20\%} & \textbf{0.83\%} & \textbf{0.29\%} & \textbf{0.77\%} & \textbf{4.50\%} & \textbf{8.33\%} & \textbf{0.92\%} & \textbf{41.07\%} \\
\bottomrule
\end{tabular}
}
\end{table*}

\subsubsection{Comparisons on MPP}
We further evaluate our method on MPP, another unified multi-physics foundation model, across four scenarios: Incompressible Naiver--Stoke, Compressible Naiver--Stoke, Diffusion--Reaction, and Shallow Water, with results reported in Table \ref{tab:mpp}.

Consistent with the VICON results, our variants generally outperform the baselines across most settings, demonstrating that the proposed framework generalizes well to unified multi-physics pretraining models. Under mild and moderate compression (ratios 0.8 and 0.6), our methods remain close to the original model on most metrics, while competing baselines already exhibit substantial degradation. At higher compression levels (ratios 0.4 and 0.2), the advantage becomes more pronounced. Several baselines deteriorate sharply, especially on more challenging scenarios such as Shallow Water and Diff React 2D. For example, at ratio 0.2 on Shallow Water, DobiSVD increases Base Loss by $27,218\%$, whereas Ours and Ours* limit the increase to $5,488\%$ and $1,299\%$, respectively. Similar trends are observed on Sobolev Loss and other metrics. The total parameter counts of the compressed models do not scale
proportionally with the target retention ratios because a substantial fraction of MPP's parameters are not subject to compression.

\begin{table*}[t]
\centering
\caption{Experimental Results on MPP.}
\label{tab:mpp}
\resizebox{\textwidth}{!}{
\begin{tabular}{c c c r r r r r r r r r r}
\toprule
\multirow{2}{*}{\textbf{$\eta$}} &
\multirow{2}{*}{\textbf{Method}} &
\multirow{2}{*}{\textbf{Size}} &
\multicolumn{4}{c}{\textbf{IncompNS}} &
\multicolumn{2}{c}{\textbf{CompNS}} &
\multicolumn{2}{c}{\textbf{DiffRe-2D}} &
\multicolumn{2}{c}{\textbf{SWE}} \\
\cmidrule(lr){4-7}
\cmidrule(lr){8-9}
\cmidrule(lr){10-11}
\cmidrule(lr){12-13}

& & &
\textbf{Base Loss $\downarrow$} &
\textbf{Sobolev $\downarrow$} &
\textbf{Div-Free $\downarrow$} &
\textbf{Vorticity $\downarrow$} &
\textbf{Base Loss $\downarrow$} &
\textbf{Sobolev $\downarrow$} &
\textbf{Base Loss $\downarrow$} &
\textbf{Sobolev $\downarrow$} &
\textbf{Base Loss $\downarrow$} &
\textbf{Sobolev $\downarrow$} \\
\midrule
1 & Origin & 407.1M & 0.00\% & 0.00\% & 0.00\% & 0.00\% & 0.00\% & 0.00\% & 0.00\% & 0.00\% & 0.00\% & 0.00\% \\
\midrule
\multirow{7}{*}{0.8} & ASVD & 366.7M & 2003\% & 72.38\% & 0.71\% & 193.6\% & 4.65\% & 0.73\% & 290.0\% & 59.07\% & 531.6\% & 75.38\% \\
 & FWSVD & 366.7M & 678.8\% & 46.03\% & 0.19\% & 97.37\% & 2.15\% & 0.65\% & 66.93\% & 23.12\% & 288.2\% & 52.56\% \\
 & SVD-LLM V2 & 366.8M & 29.15\% & 2.05\% & 0.10\% & 6.29\% & 0.86\% & 0.05\% & 25.95\% & 5.96\% & 45.31\% & 7.53\% \\
 & Dobi-SVD & 367.0M & 485.1\% & 49.66\% & 0.18\% & 86.45\% & 0.68\% & 0.06\% & 60.93\% & 24.78\% & 244.7\% & 42.97\% \\
 & SAES-SVD & 366.7M & 25.08\% & 2.00\% & \textbf{0.09\%} & 4.76\% & 0.98\% & 0.04\% & 25.98\% & 6.35\% & 56.51\% & 8.93\% \\
\hdashline
 & \textbf{Ours} & 366.8M & 28.24\% & 2.84\% & 0.12\% & 5.22\% & 0.26\% & \textbf{0.01\%} & 19.35\% & 4.28\% & 70.33\% & 8.05\% \\
 \rowcolor{gray!15} &\textbf{Ours*} & 366.8M & \textbf{16.83\%} & \textbf{1.55\%} & \textbf{0.09\%} & \textbf{3.04\%} & \textbf{0.04\%} & 0.06\% & \textbf{3.06\%} & \textbf{0.88\%} & \textbf{17.22\%} & \textbf{2.28\%} \\
\midrule
\multirow{7}{*}{0.6} & ASVD & 326.4M & 9434\% & 283.6\% & 6.28\% & 593.0\% & 64.25\% & 6.62\% & 2412\% & 210.9\% & 4836\% & 380.5\% \\
 & FWSVD & 326.4M & 3168\% & 124.6\% & 1.60\% & 262.5\% & 6.05\% & 1.68\% & 242.8\% & 61.44\% & 1311\% & 180.1\% \\
 & SVD-LLM V2 & 326.5M & 171.8\% & 10.84\% & 0.32\% & 29.86\% & 4.36\% & 0.74\% & 131.4\% & 25.14\% & 297.9\% & 42.67\% \\
 & Dobi-SVD & 326.7M & 4148\% & 207.1\% & 3.28\% & 322.8\% & 9.04\% & 1.21\% & 292.9\% & 70.01\% & 1851\% & 189.4\% \\
 & SAES-SVD & 326.4M & 122.9\% & \textbf{8.91\%} & 0.28\% & 19.62\% & 4.84\% & 0.34\% & 113.3\% & 22.36\% & 278.2\% & 39.82\% \\
\hdashline
 & \textbf{Ours} & 326.5M & 159.6\% & 15.04\% & 0.34\% & 28.12\% & \textbf{0.51\%} & \textbf{0.11\%} & 120.8\% & 19.17\% & 259.6\% & 32.76\% \\
 \rowcolor{gray!15} & \textbf{Ours*} & 326.5M & \textbf{92.62\%} & 9.14\% & \textbf{0.26\%} & \textbf{17.87\%} & 2.05\% & 0.46\% & \textbf{13.60\%} & \textbf{3.23\%} & \textbf{71.55\%} & \textbf{9.14\%} \\
\midrule
\multirow{7}{*}{0.4} & ASVD & 286.1M & 10245\% & 296.4\% & 7.04\% & 617.6\% & 72.55\% & 9.63\% & 2524\% & 216.1\% & 7345\% & 474.7\% \\
 & FWSVD & 286.1M & 9589\% & 297.9\% & 8.97\% & 584.5\% & 24.04\% & 5.18\% & 878.8\% & 153.2\% & 12771\% & 738.1\% \\
 & SVD-LLM V2 & 286.3M & 803.5\% & 39.36\% & 0.86\% & 100.5\% & 24.64\% & 3.63\% & 440.7\% & 65.71\% & 1251\% & 137.1\% \\
 & Dobi-SVD & 286.3M & 14059\% & 700.7\% & 25.47\% & 1010\% & 53.01\% & 8.07\% & 1238\% & 181.2\% & 17117\% & 729.3\% \\
 & SAES-SVD & 286.1M & 581.9\% & \textbf{31.47\%} & 0.76\% & 81.94\% & 23.37\% & 2.17\% & 377.1\% & 57.16\% & 1114\% & 122.3\% \\
\hdashline
 & \textbf{Ours} & 286.3M & 718.2\% & 56.43\% & 0.99\% & 96.92\% & \textbf{6.09\%} & \textbf{1.23\%} & 460.2\% & 59.68\% & 1188\% & 117.3\% \\
 \rowcolor{gray!15} & \textbf{Ours*} & 286.3M & \textbf{419.2\%} & 38.14\% & \textbf{0.71\%} & \textbf{70.39\%} & 12.64\% & 2.73\% & \textbf{62.80\%} & \textbf{11.25\%} & \textbf{285.3\%} & \textbf{36.63\%} \\
\midrule
\multirow{7}{*}{0.2} & ASVD & 245.8M & 12273\% & 395.7\% & 10.74\% & 806.0\% & 105.0\% & 18.32\% & 2854\% & 241.8\% & 19920\% & 849.4\% \\
 & FWSVD & 245.8M & 30072\% & 1129\% & 47.16\% & 1646\% & 109.5\% & 18.32\% & 3682\% & 422.7\% & 29702\% & 1333\% \\
 & SVD-LLM V2 & 246.0M & 3182\% & 118.5\% & \textbf{1.78\%} & 215.8\% & 124.8\% & 15.00\% & 1376\% & 138.7\% & 3203\% & 282.8\% \\
 & Dobi-SVD & 245.9M & 75128\% & 2257\% & 116.7\% & 3082\% & 371.4\% & 47.83\% & 9277\% & 713.8\% & 27218\% & 1350\% \\
 & SAES-SVD & 245.8M & 2362\% & \textbf{95.69\%} & 1.90\% & \textbf{185.6\%} & 102.6\% & 9.62\% & 1264\% & 125.8\% & 3308\% & 268.8\% \\
\hdashline
 & \textbf{Ours} & 246.0M & 2956\% & 175.7\% & 2.90\% & 245.7\% & \textbf{43.68\%} & \textbf{7.87\%} & 1514\% & 144.2\% & 5488\% & 352.4\% \\
 \rowcolor{gray!15} & \textbf{Ours*} & 246.0M & \textbf{1929\%} & 137.8\% & 2.20\% & 202.6\% & 94.22\% & 12.93\% & \textbf{317.5\%} & \textbf{42.87\%} & \textbf{1299\%} & \textbf{134.8\%} \\
\bottomrule
\end{tabular}
}
\end{table*}

\subsubsection{PDE-Specific Physical-Consistency Metrics.}
We additionally report PDE-specific physical-consistency metrics in Table~\ref{tab:physics_metrics_poseidon},
since different PDEs describe different physical scenarios.
Specifically, for CE-RM, we evaluate mass, momentum, and total-energy
conservation metrics~\ref{app:detailed_metric}. For Wave-Gauss, the wave-speed field $c(\mathbf{x})$
is time-invariant, as specified in Poseidon. We therefore report the
following constraint-based metric:
\begin{equation}
\mathcal{E}_{c}
=
\frac{1}{|\Omega|}
\sum_{\mathbf{x}\in\Omega}
\left|
\hat{c}_{t}(\mathbf{x})
-
\hat{c}_{t-\Delta t}(\mathbf{x})
\right|.
\end{equation}

\begin{table*}[tb]
\centering
\caption{PDE-specific physical-consistency metrics on Poseidon.}
\label{tab:physics_metrics_poseidon}
\resizebox{0.7\textwidth}{!}{
\begin{tabular}{c c c r r r r}
\toprule
\multirow{2}{*}{\textbf{$\eta$}} &
\multirow{2}{*}{\textbf{Method}} &
\multirow{2}{*}{\textbf{Size}} &
\multicolumn{3}{c}{\textbf{CE-RM}} &
\multicolumn{1}{c}{\textbf{Wave-Gauss}} \\
\cmidrule(lr){4-6}
\cmidrule(lr){7-7}

& & &
\textbf{Mass $\downarrow$} &
\textbf{Momentum $\downarrow$} &
\textbf{Energy $\downarrow$} &
\textbf{Speed $\downarrow$} \\
\midrule

1.0 & Origin & 628.6M &
0.00\% & 0.00\% & 0.00\% & 0.00\% \\
\midrule

\multirow{6}{*}{0.6}
& ASVD       & 376.9M & 70.62\% & 61.21\% & 67.38\% & 2093\% \\
& FWSVD      & 377.1M & 66.91\% & 59.12\% & 64.05\% & 602.6\% \\
& SVD-LLM V2 & 438.3M & 72.16\% & 55.35\% & 69.97\% & 1337\% \\
& Dobi-SVD    & 378.1M & 74.70\% & 16.49\% & 73.50\% & 1538\% \\
& SAES-SVD    & 377.1M & 45.07\% & 34.93\% & 44.02\% & 1713\% \\
\hdashline
\rowcolor{gray!15} & \textbf{Ours} & 377.5M &
\textbf{0.00\%} &
\textbf{0.00\%} &
\textbf{0.00\%} &
\textbf{0.00\%} \\
\midrule

\multirow{6}{*}{0.4}
& ASVD       & 251.4M & 71.68\% & 55.09\% & 68.56\% & 3813\% \\
& FWSVD      & 251.5M & 59.17\% & 35.82\% & 53.31\% & 967.2\% \\
& SVD-LLM V2 & 374.9M & 45.64\% & 14.00\% & 41.66\% & 3692\% \\
& Dobi-SVD    & 252.6M & 70.09\% & 24.94\% & 65.65\% & 1886\% \\
& SAES-SVD    & 251.5M & 49.01\% & 34.17\% & 46.93\% & 1947\% \\
\hdashline
\rowcolor{gray!15} & \textbf{Ours} & 251.9M &
\textbf{0.00\%} &
\textbf{0.01\%} &
\textbf{0.01\%} &
\textbf{0.26\%} \\
\midrule

\multirow{6}{*}{0.2}
& ASVD       & 125.8M & 58.53\% & 57.60\% & 53.09\% & 3347\% \\
& FWSVD      & 125.9M & 48.62\% & 93.66\% & 43.75\% & 2457\% \\
& SVD-LLM V2 & 353.2M & 35.33\% & 35.92\% & 34.41\% & 4677\% \\
& Dobi-SVD    & 127.0M & 69.53\% & 36.65\% & 62.92\% & 3257\% \\
& SAES-SVD    & 125.9M & 55.31\% & 33.72\% & 51.94\% & 2159\% \\
\hdashline
\rowcolor{gray!15} & \textbf{Ours} & 126.3M &
\textbf{0.72\%} &
\textbf{0.65\%} &
\textbf{0.86\%} &
\textbf{96.47\%} \\

\bottomrule
\end{tabular}
}
\end{table*}

\subsubsection{Main Paper Full Results on Poseidon and VICON}
Also, the results of VICON, and NS-PwC, CE-RM and Wave-Gaussian (containing 0.8 target retention ratio) on Poseidon are shown in Table \ref{tab: whole_vicon} and Table \ref{tab: whole_poseidon} respectively.

\begin{table*}[t]
\centering
\caption{Experimental Results on Poseidon}
\label{tab: whole_poseidon}
\resizebox{\textwidth}{!}{
\begin{tabular}{c c c r r r r r r r r}
\toprule
\multirow{2}{*}{\textbf{$\eta$}} &
\multirow{2}{*}{\textbf{Method}} &
\multirow{2}{*}{\textbf{Size}} &
\multicolumn{4}{c}{\textbf{NS-PwC}} &
\multicolumn{2}{c}{\textbf{CE-RM}} &
\multicolumn{2}{c}{\textbf{Wave-Gauss}} \\
\cmidrule(lr){4-7}
\cmidrule(lr){8-9}
\cmidrule(lr){10-11}

& & &
\textbf{Base Loss $\downarrow$} &
\textbf{Sobolev $\downarrow$} &
\textbf{Div-Free $\downarrow$} &
\textbf{Vorticity $\downarrow$} &
\textbf{Base Loss $\downarrow$} &
\textbf{Sobolev $\downarrow$} &
\textbf{Base Loss $\downarrow$} &
\textbf{Sobolev $\downarrow$} \\
\midrule
1 & Origin & 628.6M & 0.00\% & 0.00\% & 0.00\% & 0.00\% & 0.00\% & 0.00\% & 0.00\% & 0.00\% \\
\midrule
\multirow{6}{*}{0.8} & ASVD & 502.9M & 2000\% & 884.6\% & 498.7\% & 1486\% & 9.01\% & 1.79\% & 81.85\% & 1090\% \\
 & FWSVD & 502.7M & 6930\% & 2276\% & 1345\% & 3657\% & 20.72\% & 0.50\% & 559.6\% & 1779\% \\
 & SVD-LLM V2 & 524.2M & 6851\% & 1781\% & 1102\% & 3602\% & 22.81\% & 4.34\% & 556.6\% & 1256\% \\
 & Dobi-SVD & 503.7M & 7961\% & 2163\% & 1283\% & 3790\% & 34.17\% & 5.31\% & 632.9\% & 2202\% \\
 & SAES-SVD & 502.7M & 2616\% & 1653\% & 1464\% & 1670\% & 1.07\% & 1.13\% & 229.9\% & 2438\% \\
\hdashline
\rowcolor{gray!15} & \textbf{Ours} & 503.0M & \textbf{0.00\%} & \textbf{0.00\%} & \textbf{0.00\%} & \textbf{0.00\%} & \textbf{0.00\%} & \textbf{0.00\%} & \textbf{0.00\%} & \textbf{0.00\%} \\
\midrule
\multirow{6}{*}{0.6} & ASVD & 376.9M & 4496\% & 2680\% & 2017\% & 2805\% & 25.39\% & 6.61\% & 352.1\% & 6202\% \\
 & FWSVD & 377.1M & 7124\% & 2977\% & 1700\% & 3727\% & 23.24\% & 8.51\% & 569.7\% & 2415\% \\
 & SVD-LLM V2 & 438.3M & 7038\% & 2449\% & 1482\% & 3651\% & 26.78\% & 9.31\% & 576.4\% & 4982\% \\
 & Dobi-SVD & 378.1M & 7468\% & 3036\% & 2137\% & 3719\% & 35.70\% & 24.58\% & 658.6\% & 9210\% \\
 & SAES-SVD & 377.1M & 2882\% & 1763\% & 1572\% & 1863\% & 2.19\% & 2.53\% & 225.8\% & 3354\% \\
\hdashline
\rowcolor{gray!15} & \textbf{Ours} & 377.5M & \textbf{0.00\%} & \textbf{0.01\%} & \textbf{0.00\%} & \textbf{0.01\%} & \textbf{0.00\%} & \textbf{0.00\%} & \textbf{0.00\%} & \textbf{0.01\%} \\
\midrule
\multirow{6}{*}{0.4} & ASVD & 251.4M & 5169\% & 4364\% & 2773\% & 3311\% & 34.10\% & 15.79\% & 475.4\% & 14572\% \\
 & FWSVD & 251.5M & 7846\% & 5080\% & 2449\% & 3961\% & 28.33\% & 28.64\% & 588.1\% & 3542\% \\
 & SVD-LLM V2 & 374.9M & 8741\% & 6051\% & 4530\% & 4131\% & 35.87\% & 54.99\% & 656.0\% & 19627\% \\
 & Dobi-SVD & 252.6M & 8481\% & 5206\% & 2709\% & 4048\% & 39.27\% & 25.91\% & 670.4\% & 10497\% \\
 & SAES-SVD & 251.5M & 3062\% & 1868\% & 1632\% & 1967\% & 3.46\% & 6.18\% & 249.3\% & 4020\% \\
\hdashline
\rowcolor{gray!15} & \textbf{Ours} & 251.9M & \textbf{0.26\%} & \textbf{0.08\%} & \textbf{0.03\%} & \textbf{0.17\%} & \textbf{0.01\%} & \textbf{0.05\%} & \textbf{0.04\%} & \textbf{0.40\%} \\
\midrule
\multirow{6}{*}{0.2} & ASVD & 125.8M & 6725\% & 6183\% & 3563\% & 4167\% & 41.02\% & 80.05\% & 525.0\% & 13007\% \\
 & FWSVD & 125.9M & 8622\% & 7650\% & 3770\% & 4276\% & 40.57\% & 97.34\% & 655.1\% & 11452\% \\
 & SVD-LLM V2 & 353.2M & 9529\% & 6011\% & 4183\% & 4434\% & 61.44\% & 66.29\% & 1122\% & 19865\% \\
 & Dobi-SVD & 127.0M & 8774\% & 10090\% & 4999\% & 4440\% & 44.78\% & 29.72\% & 728.6\% & 20124\% \\
 & SAES-SVD & 125.9M & 3726\% & 2429\% & 2199\% & 2426\% & 6.25\% & 9.83\% & 366.9\% & 5873\% \\
\hdashline
\rowcolor{gray!15} & \textbf{Ours} & 126.3M & \textbf{2.64\%} & \textbf{1.79\%} & \textbf{0.52\%} & \textbf{1.35\%} & \textbf{0.10\%} & \textbf{1.01\%} & \textbf{0.70\%} & \textbf{151.9\%} \\
\bottomrule
\end{tabular}
}
\end{table*}
\begin{table*}[t]
\centering
\caption{Experimental Results on VICON.}
\label{tab: whole_vicon}
\resizebox{\textwidth}{!}{
\begin{tabular}{c c c r r r r r r r r}
\toprule
\multirow{2}{*}{\textbf{$\eta$}} &
\multirow{2}{*}{\textbf{Method}} &
\multirow{2}{*}{\textbf{Size}} &
\multicolumn{4}{c}{\textbf{NS-2D}} &
\multicolumn{2}{c}{\textbf{Comp-2D}} &
\multicolumn{2}{c}{\textbf{Euler-2D}} \\
\cmidrule(lr){4-7}
\cmidrule(lr){8-9}
\cmidrule(lr){10-11}

& & &
\textbf{Base Loss $\downarrow$} &
\textbf{Sobolev $\downarrow$} &
\textbf{Div-Free $\downarrow$} &
\textbf{Vorticity $\downarrow$} &
\textbf{Base Loss $\downarrow$} &
\textbf{Sobolev $\downarrow$} &
\textbf{Base Loss $\downarrow$} &
\textbf{Sobolev $\downarrow$} \\
\midrule
1 & Origin & 87.8M & 0.00\% & 0.00\% & 0.00\% & 0.00\% & 0.00\% & 0.00\% & 0.00\% & 0.00\% \\
\midrule
\multirow{7}{*}{0.8} & ASVD & 70.1M & 1922\% & 48.48\% & 18.58\% & 184.9\% & 14247\% & 317.2\% & 1219\% & 30.49\% \\
 & FWSVD & 70.3M & 1011\% & 253.7\% & 14.36\% & 302.3\% & 2859\% & 748.3\% & 276.4\% & 12.23\% \\
 & SVD-LLM V2 & 70.2M & 16.64\% & 2.92\% & 1.90\% & 11.08\% & 65.11\% & 22.19\% & 16.27\% & \textbf{1.00\%} \\
 & Dobi-SVD & 69.1M & 1762\% & 108.3\% & 16.06\% & 147.0\% & 277.9\% & 230.4\% & 66.02\% & 21.04\% \\
 & SAES-SVD & 70.3M & 15.15\% & 3.44\% & 1.48\% & 9.90\% & 52.18\% & 19.86\% & 10.04\% & 8.99\% \\
\hdashline
 & \textbf{Ours} & 70.3M & 4.61\% & 1.07\% & 0.51\% & 1.92\% & 6.17\% & 5.40\% & 4.14\% & 29.26\% \\
 \rowcolor{gray!15} & \textbf{Ours*} & 70.3M & \textbf{0.58\%} & \textbf{0.35\%} & \textbf{0.32\%} & \textbf{0.02\%} & \textbf{0.75\%} & \textbf{0.85\%} & \textbf{2.55\%} & 29.00\% \\
\midrule
\multirow{7}{*}{0.6} & ASVD & 52.4M & 10140\% & 528.7\% & 7.01\% & 476.1\% & 41318\% & 1394\% & 2692\% & 97.36\% \\
 & FWSVD & 52.7M & 1861\% & 302.3\% & 2.23\% & 357.5\% & 7885\% & 947.6\% & 805.1\% & 57.72\% \\
 & SVD-LLM V2 & 52.7M & 73.76\% & 11.74\% & 5.30\% & 43.87\% & 260.0\% & 60.00\% & 47.26\% & \textbf{3.31\%} \\
 & Dobi-SVD & 52.9M & 1968\% & 109.3\% & 24.08\% & 163.6\% & 757.4\% & 260.1\% & 128.6\% & 18.49\% \\
 & SAES-SVD & 52.7M & 49.63\% & 7.06\% & 3.92\% & 26.01\% & 144.7\% & 33.70\% & 36.66\% & 25.99\% \\
\hdashline
 & \textbf{Ours} & 52.8M & 2.07\% & 1.22\% & \textbf{0.21\%} & 3.08\% & \textbf{16.54\%} & \textbf{3.82\%} & \textbf{3.08\%} & 28.35\% \\
 \rowcolor{gray!15} & \textbf{Ours*} & 52.8M & \textbf{0.51\%} & \textbf{0.02\%} & 0.71\% & \textbf{0.06\%} & 18.05\% & 10.86\% & 8.52\% & 28.89\% \\
\midrule
\multirow{7}{*}{0.4} & ASVD & 33.5M & 13064\% & 532.5\% & 11.34\% & 490.4\% & 46290\% & 1382\% & 2865\% & 94.70\% \\
 & FWSVD & 35.2M & 4862\% & 349.7\% & 3.24\% & 437.5\% & 22440\% & 1155\% & 2010\% & 69.01\% \\
 & SVD-LLM V2 & 35.7M & 689.5\% & 32.06\% & 10.96\% & 136.1\% & 2152\% & 163.9\% & 265.6\% & 16.69\% \\
 & Dobi-SVD & 37.0M & 2607\% & 113.3\% & 34.70\% & 209.6\% & 1795\% & 285.9\% & 278.3\% & \textbf{13.58\%} \\
 & SAES-SVD & 35.2M & 156.3\% & 18.10\% & 8.68\% & 68.40\% & 609.3\% & 74.10\% & 115.5\% & 20.65\% \\
\hdashline
 & \textbf{Ours} & 35.3M & 24.98\% & 4.23\% & \textbf{0.85\%} & 11.32\% & 127.4\% & \textbf{26.40\%} & 27.05\% & 26.91\% \\
 \rowcolor{gray!15} & \textbf{Ours*} & 35.3M & \textbf{18.92\%} & \textbf{2.90\%} & 1.30\% & \textbf{7.26\%} & \textbf{81.80\%} & 29.87\% & \textbf{26.11\%} & 27.55\% \\
\midrule
\multirow{7}{*}{0.2} & ASVD & 16.5M & 15583\% & 535.6\% & 12.51\% & 484.5\% & 48919\% & 1392\% & 3016\% & 94.46\% \\
 & FWSVD & 17.7M & 11265\% & 295.8\% & 29.67\% & 462.3\% & 43908\% & 1232\% & 2769\% & 66.10\% \\
 & SVD-LLM V2 & 18.3M & 2245\% & 67.19\% & 40.38\% & 290.2\% & 16822\% & 399.9\% & 1230\% & 34.89\% \\
 & Dobi-SVD & 20.2M & 3705\% & 118.8\% & 46.52\% & 262.2\% & 4929\% & 358.1\% & 630.2\% & \textbf{7.02\%} \\
 & SAES-SVD & 17.7M & 651.3\% & 37.96\% & 16.22\% & 165.2\% & 2952\% & 169.7\% & 432.3\% & 12.12\% \\
\hdashline
 & \textbf{Ours} & 17.8M & 291.1\% & 23.19\% & \textbf{7.87\%} & 76.27\% & 1051\% & \textbf{99.28\%} & 291.8\% & 16.10\% \\
\rowcolor{gray!15} & \textbf{Ours*} & 17.8M & \textbf{239.7\%} & \textbf{20.17\%} & 8.07\% & \textbf{66.31\%} & \textbf{927.2\%} & 116.8\% & \textbf{248.3\%} & 17.75\% \\
\bottomrule
\end{tabular}
}
\end{table*}

\subsubsection{Multi-Seed Robustness}
\label{app:multi_seed}

To assess the robustness of SAFE-SVD to randomness, we conduct
additional runs for Poseidon , VICON, and MPP under multiple random
seeds, which are shown in Table~\ref{tab:poseidon_multiseed}, Table~\ref{tab:vicon_multiseed}, and Table~\ref{tab:mpp_multiseed} respectively. For each setting, we report statistics over 11 runs in total,
including the run reported in the main paper. Results are presented
as mean $\pm$ standard deviation. For Poseidon, we retain six decimal
places because several relative changes are close to zero.

For Poseidon, the main paper default Poseidon results are consistent with the multi-seed results
across all evaluated settings. For VICON and MPP, we observe a few
small deviations from the originally reported values, e.g.,
IncompNS Sobolev loss at retention ratio 0.2 and Div-Free error at
0.8 on MPP, and NS-2D Base Loss and Vorticity at 0.8 on VICON.

\begin{table*}[tb]
\centering
\caption{Multi-seed results on Poseidon.}
\label{tab:poseidon_multiseed}
\resizebox{\textwidth}{!}{
\begin{tabular}{c c c r r r r r r r r}
\toprule
\multirow{2}{*}{\textbf{$\eta$}} &
\multirow{2}{*}{\textbf{Method}} &
\multirow{2}{*}{\textbf{Size}} &
\multicolumn{4}{c}{\textbf{NS-PwC}} &
\multicolumn{2}{c}{\textbf{CE-RM}} &
\multicolumn{2}{c}{\textbf{Wave-Gauss}} \\
\cmidrule(lr){4-7}
\cmidrule(lr){8-9}
\cmidrule(lr){10-11}

& & &
\textbf{Base Loss $\downarrow$} &
\textbf{Sobolev $\downarrow$} &
\textbf{Div-Free $\downarrow$} &
\textbf{Vorticity $\downarrow$} &
\textbf{Base Loss $\downarrow$} &
\textbf{Sobolev $\downarrow$} &
\textbf{Base Loss $\downarrow$} &
\textbf{Sobolev $\downarrow$} \\
\midrule

0.8 & Ours & 503.0M &
$0.000325\% \pm 0.000022\%$ &
$0.000056\% \pm 0.000005\%$ &
$0.000079\% \pm 0.000006\%$ &
$0.000063\% \pm 0.000009\%$ &
$0.000009\% \pm 0.000000\%$ &
$0.000024\% \pm 0.000001\%$ &
$0.000130\% \pm 0.000009\%$ &
$0.000028\% \pm 0.000001\%$ \\

0.6 & Ours & 377.5M &
$0.004377\% \pm 0.000993\%$ &
$0.010893\% \pm 0.003671\%$ &
$0.002497\% \pm 0.000175\%$ &
$0.011064\% \pm 0.000611\%$ &
$0.003824\% \pm 0.000264\%$ &
$0.002570\% \pm 0.000152\%$ &
$0.002638\% \pm 0.000182\%$ &
$0.006615\% \pm 0.000309\%$ \\

0.4 & Ours & 251.9M &
$0.272097\% \pm 0.018793\%$ &
$0.082933\% \pm 0.009439\%$ &
$0.031330\% \pm 0.001561\%$ &
$0.167792\% \pm 0.008715\%$ &
$0.012852\% \pm 0.000888\%$ &
$0.046100\% \pm 0.002722\%$ &
$0.042955\% \pm 0.002967\%$ &
$0.389128\% \pm 0.018195\%$ \\

0.2 & Ours & 126.3M &
$2.630446\% \pm 0.173058\%$ &
$1.817209\% \pm 0.179115\%$ &
$0.491240\% \pm 0.034353\%$ &
$1.324470\% \pm 0.091498\%$ &
$0.102373\% \pm 0.006863\%$ &
$0.992629\% \pm 0.079074\%$ &
$0.696032\% \pm 0.048073\%$ &
$147.252537\% \pm 7.164102\%$ \\

\bottomrule
\end{tabular}
}
\end{table*}
\begin{table*}[tb]
\centering
\caption{Multi-seed results on VICON.}
\label{tab:vicon_multiseed}
\resizebox{\textwidth}{!}{
\begin{tabular}{c c c r r r r r r r r}
\toprule
\multirow{2}{*}{\textbf{$\eta$}} &
\multirow{2}{*}{\textbf{Method}} &
\multirow{2}{*}{\textbf{Size}} &
\multicolumn{4}{c}{\textbf{NS-2D}} &
\multicolumn{2}{c}{\textbf{Comp-2D}} &
\multicolumn{2}{c}{\textbf{Euler-2D}} \\
\cmidrule(lr){4-7}
\cmidrule(lr){8-9}
\cmidrule(lr){10-11}

& & &
\textbf{Base Loss $\downarrow$} &
\textbf{Sobolev $\downarrow$} &
\textbf{Div-Free $\downarrow$} &
\textbf{Vorticity $\downarrow$} &
\textbf{Base Loss $\downarrow$} &
\textbf{Sobolev $\downarrow$} &
\textbf{Base Loss $\downarrow$} &
\textbf{Sobolev $\downarrow$} \\
\midrule

0.8 & Ours$^\ast$ & 70.3M &
$2.32\% \pm 1.50\%$ &
$0.61\% \pm 0.44\%$ &
$0.17\% \pm 0.09\%$ &
$0.55\% \pm 0.37\%$ &
$2.20\% \pm 1.34\%$ &
$0.91\% \pm 0.48\%$ &
$3.15\% \pm 2.50\%$ &
$18.37\% \pm 12.94\%$ \\

0.6 & Ours$^\ast$ & 52.8M &
$2.09\% \pm 1.58\%$ &
$0.38\% \pm 0.47\%$ &
$0.37\% \pm 0.14\%$ &
$1.21\% \pm 0.54\%$ &
$17.10\% \pm 3.59\%$ &
$7.74\% \pm 1.69\%$ &
$7.19\% \pm 3.63\%$ &
$15.51\% \pm 13.43\%$ \\

0.4 & Ours$^\ast$ & 35.3M &
$20.64\% \pm 1.62\%$ &
$3.23\% \pm 0.53\%$ &
$1.01\% \pm 0.14\%$ &
$8.47\% \pm 0.62\%$ &
$73.42\% \pm 7.41\%$ &
$25.93\% \pm 2.38\%$ &
$28.07\% \pm 3.71\%$ &
$16.69\% \pm 12.52\%$ \\

0.2 & Ours$^\ast$ & 17.8M &
$239.0\% \pm 6.25\%$ &
$20.41\% \pm 0.61\%$ &
$7.43\% \pm 0.31\%$ &
$66.07\% \pm 1.45\%$ &
$984.0\% \pm 79.79\%$ &
$104.2\% \pm 6.49\%$ &
$257.2\% \pm 10.65\%$ &
$14.34\% \pm 4.83\%$ \\

\bottomrule
\end{tabular}
}
\end{table*}
\begin{table*}[tb]
\centering
\caption{Multi-seed results on MPP.}
\label{tab:mpp_multiseed}
\resizebox{\textwidth}{!}{
\begin{tabular}{c c c r r r r r r r r r r}
\toprule
\multirow{2}{*}{\textbf{$\eta$}} &
\multirow{2}{*}{\textbf{Method}} &
\multirow{2}{*}{\textbf{Size}} &
\multicolumn{4}{c}{\textbf{IncompNS}} &
\multicolumn{2}{c}{\textbf{CompNS}} &
\multicolumn{2}{c}{\textbf{DiffRe-2D}} &
\multicolumn{2}{c}{\textbf{SWE}} \\
\cmidrule(lr){4-7}
\cmidrule(lr){8-9}
\cmidrule(lr){10-11}
\cmidrule(lr){12-13}

& & &
\textbf{Base Loss $\downarrow$} &
\textbf{Sobolev $\downarrow$} &
\textbf{Div-Free $\downarrow$} &
\textbf{Vorticity $\downarrow$} &
\textbf{Base Loss $\downarrow$} &
\textbf{Sobolev $\downarrow$} &
\textbf{Base Loss $\downarrow$} &
\textbf{Sobolev $\downarrow$} &
\textbf{Base Loss $\downarrow$} &
\textbf{Sobolev $\downarrow$} \\
\midrule

0.8 & Ours$^\ast$ & 366.8M &
$15.45\% \pm 1.91\%$ &
$1.54\% \pm 0.20\%$ &
$0.06\% \pm 0.02\%$ &
$2.81\% \pm 0.32\%$ &
$0.15\% \pm 0.14\%$ &
$0.08\% \pm 0.02\%$ &
$3.73\% \pm 0.69\%$ &
$0.89\% \pm 0.08\%$ &
$18.78\% \pm 2.08\%$ &
$2.38\% \pm 0.30\%$ \\

0.6 & Ours$^\ast$ & 326.5M &
$90.86\% \pm 4.33\%$ &
$8.88\% \pm 0.61\%$ &
$0.22\% \pm 0.07\%$ &
$17.06\% \pm 0.99\%$ &
$1.89\% \pm 0.46\%$ &
$0.55\% \pm 0.11\%$ &
$15.70\% \pm 2.19\%$ &
$3.31\% \pm 0.21\%$ &
$71.20\% \pm 9.84\%$ &
$9.76\% \pm 0.98\%$ \\

0.4 & Ours$^\ast$ & 286.3M &
$407.5\% \pm 10.54\%$ &
$36.07\% \pm 2.19\%$ &
$0.64\% \pm 0.14\%$ &
$66.54\% \pm 4.12\%$ &
$13.20\% \pm 1.89\%$ &
$2.84\% \pm 0.30\%$ &
$65.43\% \pm 6.50\%$ &
$11.59\% \pm 0.68\%$ &
$277.3\% \pm 24.42\%$ &
$36.00\% \pm 1.83\%$ \\

0.2 & Ours$^\ast$ & 246.0M &
$1882\% \pm 104.6\%$ &
$129.9\% \pm 7.17\%$ &
$2.05\% \pm 0.22\%$ &
$197.5\% \pm 9.93\%$ &
$98.35\% \pm 10.17\%$ &
$12.74\% \pm 0.70\%$ &
$318.1\% \pm 25.30\%$ &
$43.50\% \pm 1.71\%$ &
$1246\% \pm 89.80\%$ &
$131.5\% \pm 5.79\%$ \\

\bottomrule
\end{tabular}
}
\end{table*}

\subsubsection{$l_1$ vs.\ $l_2$ Sobolev Loss}
\label{app:l1_l2}

Our ablation study in Section~\ref{main:abalation_study} compares the $l_1$ and $l_2$
Sobolev losses on Poseidon NS-BB, where the $l_1$ variant consistently
achieves better performance. Based on this observation, we adopt the
$l_1$ Sobolev loss as the default setting in the main experiments.

To examine whether this observation generalizes across physics
foundation models, we further evaluate SAFE-SVD using the $l_2$
Sobolev loss on Poseidon, VICON, and MPP. The complete results are
reported in Table~\ref{tab:l2_poseidon}, \ref{tab:l2_vicon} and \ref{tab:l2_mpp}.
\begin{table*}[tb]
\centering
\caption{$l_2$ Sobolev loss results on Poseidon.}
\label{tab:l2_poseidon}
\resizebox{\textwidth}{!}{
\begin{tabular}{c c c r r r r r r r r}
\toprule
\textbf{$\eta$} &
\textbf{Method} &
\textbf{Size} &
\multicolumn{4}{c}{\textbf{NS-PwC}} &
\multicolumn{2}{c}{\textbf{CE-RM}} &
\multicolumn{2}{c}{\textbf{Wave-Gauss}} \\
\cmidrule(lr){4-7}
\cmidrule(lr){8-9}
\cmidrule(lr){10-11}

& & &
\textbf{Base Loss $\downarrow$} &
\textbf{Sobolev $\downarrow$} &
\textbf{Div-Free $\downarrow$} &
\textbf{Vorticity $\downarrow$} &
\textbf{Base Loss $\downarrow$} &
\textbf{Sobolev $\downarrow$} &
\textbf{Base Loss $\downarrow$} &
\textbf{Sobolev $\downarrow$} \\
\midrule

1.0 & Origin & 628.6M &
0.00\% & 0.00\% & 0.00\% & 0.00\% &
0.00\% & 0.00\% &
0.00\% & 0.00\% \\

0.8 & Ours & 503.0M &
0.02\% & 0.00\% & 0.00\% & 0.00\% &
0.00\% & 0.00\% &
2.55\% & 0.34\% \\

0.6 & Ours & 377.5M &
0.01\% & 0.01\% & 0.01\% & 0.01\% &
0.00\% & 0.01\% &
2.27\% & 157.5\% \\

0.4 & Ours & 251.9M &
0.73\% & 3.39\% & 0.83\% & 0.34\% &
0.00\% & 0.70\% &
2.09\% & 1923\% \\

0.2 & Ours & 126.3M &
58.04\% & 224.2\% & 119.0\% & 30.76\% &
1.04\% & 18.54\% &
33.04\% & 10547\% \\

\bottomrule
\end{tabular}
}
\end{table*}
\begin{table*}[tb]
\centering
\caption{$l_2$ Sobolev loss results on VICON.}
\label{tab:l2_vicon}
\resizebox{\textwidth}{!}{
\begin{tabular}{c c c r r r r r r r r}
\toprule
\textbf{$\eta$} &
\textbf{Method} &
\textbf{Size} &
\multicolumn{4}{c}{\textbf{NS-2D}} &
\multicolumn{2}{c}{\textbf{Comp-2D}} &
\multicolumn{2}{c}{\textbf{Euler-2D}} \\
\cmidrule(lr){4-7}
\cmidrule(lr){8-9}
\cmidrule(lr){10-11}

& & &
\textbf{Base Loss $\downarrow$} &
\textbf{Sobolev $\downarrow$} &
\textbf{Div-Free $\downarrow$} &
\textbf{Vorticity $\downarrow$} &
\textbf{Base Loss $\downarrow$} &
\textbf{Sobolev $\downarrow$} &
\textbf{Base Loss $\downarrow$} &
\textbf{Sobolev $\downarrow$} \\
\midrule

1.0 & Origin & 87.8M &
0.00\% & 0.00\% & 0.00\% & 0.00\% &
0.00\% & 0.00\% &
0.00\% & 0.00\% \\

0.8 & Ours* & 70.3M &
0.42\% & 0.17\% & 0.05\% & 0.21\% &
1.46\% & 0.81\% &
0.20\% & 0.00\% \\

0.6 & Ours* & 52.8M &
7.83\% & 2.32\% & 0.43\% & 4.56\% &
16.24\% & 11.73\% &
4.59\% & 0.79\% \\

0.4 & Ours* & 35.3M &
34.36\% & 7.50\% & 1.93\% & 18.29\% &
97.59\% & 35.44\% &
27.54\% & 3.61\% \\

0.2 & Ours* & 17.8M &
330.1\% & 28.92\% & 11.13\% & 107.1\% &
782.9\% & 126.0\% &
274.8\% & 18.20\% \\

\bottomrule
\end{tabular}
}
\end{table*}
\begin{table*}[tb]
\centering
\caption{$l_2$ Sobolev loss results on MPP.}
\label{tab:l2_mpp}
\resizebox{\textwidth}{!}{
\begin{tabular}{c c r r r r r r r r r r r}
\toprule
\textbf{$\eta$} &
\textbf{Method} &
\textbf{Size} &
\multicolumn{4}{c}{\textbf{IncompNS}} &
\multicolumn{2}{c}{\textbf{CompNS}} &
\multicolumn{2}{c}{\textbf{DiffRe-2D}} &
\multicolumn{2}{c}{\textbf{SWE}} \\
\cmidrule(lr){4-7}
\cmidrule(lr){8-9}
\cmidrule(lr){10-11}
\cmidrule(lr){12-13}

& & &
\textbf{Base Loss $\downarrow$} &
\textbf{Sobolev $\downarrow$} &
\textbf{Div-Free $\downarrow$} &
\textbf{Vorticity $\downarrow$} &
\textbf{Base Loss $\downarrow$} &
\textbf{Sobolev $\downarrow$} &
\textbf{Base Loss $\downarrow$} &
\textbf{Sobolev $\downarrow$} &
\textbf{Base Loss $\downarrow$} &
\textbf{Sobolev $\downarrow$} \\
\midrule

1.0 & Origin & 407.1M &
0.00\% & 0.00\% & 0.00\% & 0.00\% &
0.00\% & 0.00\% &
0.00\% & 0.00\% &
0.00\% & 0.00\% \\

0.8 & Ours* & 366.8M &
17.81\% & 1.78\% & 0.09\% & 3.23\% &
0.24\% & 0.02\% &
3.72\% & 0.82\% &
21.57\% & 2.54\% \\

0.6 & Ours* & 326.5M &
100.8\% & 10.20\% & 0.28\% & 19.33\% &
3.55\% & 0.40\% &
18.42\% & 3.28\% &
82.32\% & 10.09\% \\

0.4 & Ours* & 286.3M &
422.2\% & 38.31\% & 0.68\% & 67.53\% &
18.84\% & 2.38\% &
78.80\% & 12.04\% &
289.1\% & 37.13\% \\

0.2 & Ours* & 246.0M &
1851\% & 124.3\% & 1.90\% & 183.6\% &
152.3\% & 14.01\% &
354.9\% & 45.89\% &
1215\% & 131.4\% \\

\bottomrule
\end{tabular}
}
\end{table*}
The additional results show that the relative performance of
$l_1$ and $l_2$ Sobolev losses is model- and setting-dependent.
On Poseidon, the $l_1$ loss generally provides substantially better
compression robustness, consistent with the ablation results in
Section~4.1. However, this trend does not hold uniformly for VICON
and MPP: the $l_2$ variant performs better on some metrics and
retention ratios, while the $l_1$ variant performs better on others.
Therefore, our ablation supports $l_1$ as an effective default for
Poseidon, rather than establishing a universal superiority of
$l_1$ over $l_2$ across all PFMs. For consistency, we retain the
$l_1$ configuration selected from the original Poseidon ablation
as the default across the main experiments, rather than selecting
the norm separately for each model or dataset.

\subsection{Qualitative Results}

\paragraph{CE-RM}

\begin{figure}[htbp]
    \centering
    \begin{subfigure}{\textwidth}
        \includegraphics[width=\textwidth]{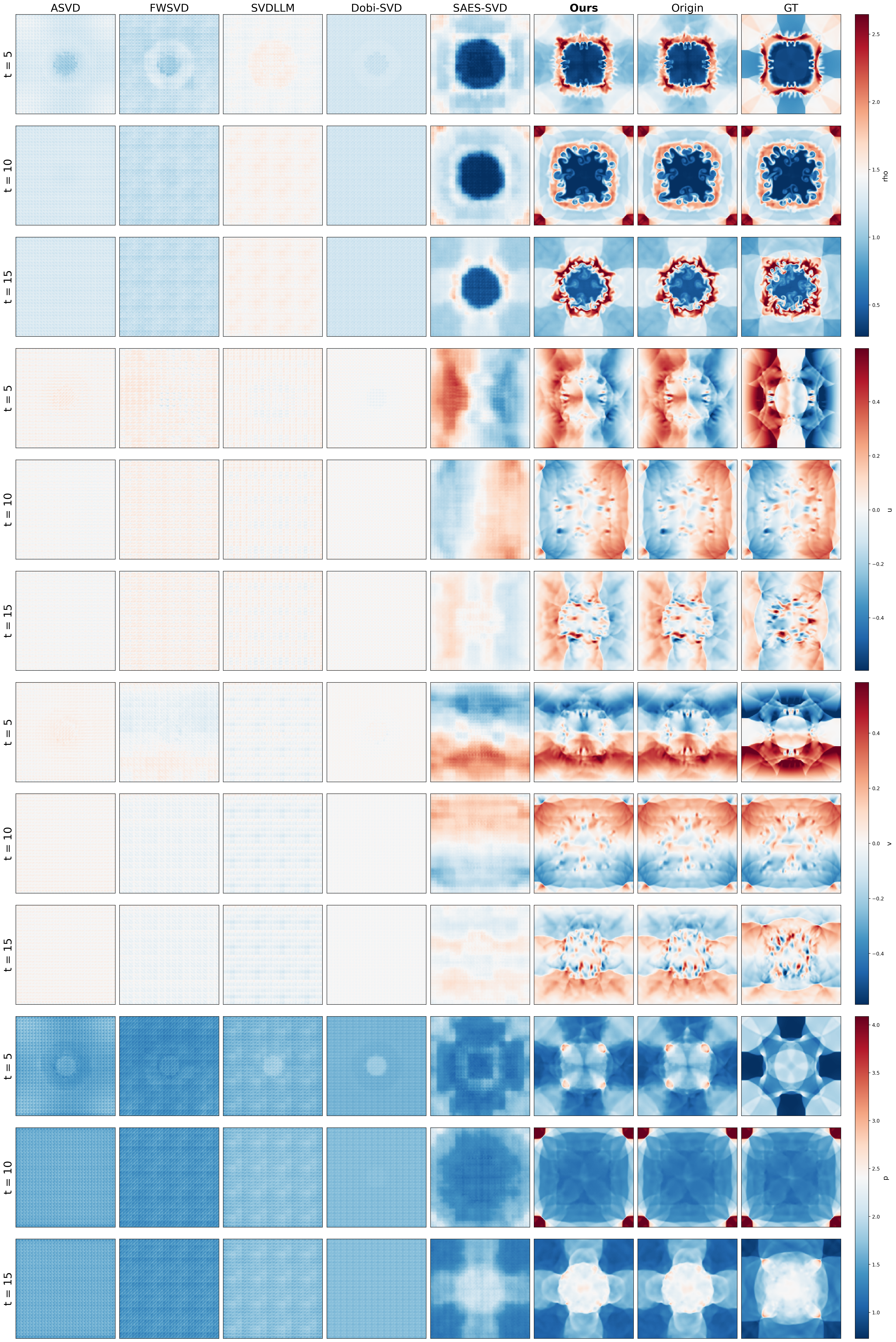}
        \label{fig: quanli_CE_RM}
    \end{subfigure}
    \caption{Predicted physical fields ($\rho, u, v, p$) on the CE-RM dataset at
    target retention ratio 0.2 over three rollout horizons ($t=5, 10, 15$).}
    \label{figs:quali_CE_RM}
\end{figure}

\cref{figs:quali_CE_RM} compares predicted physical fields ($\rho, u, v, p$) on
the CE-RM (Compressible Euler--Richtmyer--Meshkov) dataset. This dataset is characterized by high-Mach shock waves and complex interfacial instabilities that produce sharp discontinuities and intricate mushroom filament structures in the density field.

ASVD, FWSVD, SVD-LLM V2, and Dobi-SVD suffer from near-complete information
collapse; their predictions are largely featureless, failing to capture even the
most basic macroscopic structures.  SAES-SVD recovers the global symmetry of the
expanding flow but significantly blurs the sharp discontinuities and lacks the
high-frequency components needed to resolve fine-scale shock reflections.  Our
method, by contrast, demonstrates remarkable visual consistency with the
original model across all physical variables, preserving the sharp shock fronts,
the Richtmyer--Meshkov mushroom structures in the density field, and the complex
wave interference patterns in the velocity and pressure fields even at
$t{=}15$.


\section{Experiment Setting Details}
\label{app:dataset_config}
\subsection{Calibration Dataset}
Our calibration dataset consists of 13,440 samples ($210\text{ samples}/\text{per trajectory} \times 64$ trajectories) for the Incompressible Navier-Stokes and Compressible Euler tasks, and 7,680 samples ($120\text{ samples}/\text{trajectory} \times 64$ trajectories) for the Wave Equation. The sample counts per trajectory are derived from the summation of time steps up to the final sampled time:
\begin{equation}
    N = \sum_{i=1}^{T_{final}} i,
\end{equation}
where $T_{final}=20$ for Navier-Stokes and Euler tasks ($N=210$), and $T_{final}=15$ for the Wave Equation ($N=120$). Furthermore, we include 2,048 samples for VICON and 16 samples for MPP per physical scenario. These configurations were selected to provide an optimal trade-off between the preservation of model fidelity and overall compression efficiency.

\paragraph{Calibration-Size Trade-off.}
We further study the effect of calibration-set size on SAFE-SVD
using Poseidon NS-PwC at a target retention ratio of 0.2.
Table~\ref{tab:calib_tradeoff} reports both compression cost and
prediction quality for different numbers of calibration trajectories.

\begin{table*}[tb]
\centering
\caption{Accuracy--compute trade-off on Poseidon NS-PwC at a target parameter retention ratio of 0.2.}
\label{tab:calib_tradeoff}
\resizebox{\textwidth}{!}{
\begin{tabular}{c c r r r r r r r}
\toprule
\textbf{Method} &
\textbf{Calib. traj.} &
\textbf{Rank selection (min)} &
\textbf{Compression (min)} &
\textbf{Total (min)} &
\textbf{GPU alloc. peak (GB)} &
\textbf{Base Loss $\downarrow$} &
\textbf{Div-Free $\downarrow$} &
\textbf{Vorticity $\downarrow$} \\
\midrule

\multirow{5}{*}{Ours}
& 4  & 2.82  & 34.40  & 37.22  & 15.46 & 3.93\% & 0.68\% & 2.93\% \\
& 8  & 4.62  & 69.94  & 74.55  & 15.46 & 3.55\% & 0.83\% & 2.36\% \\
& 16 & 7.59  & 131.06 & 138.65 & 16.83 & 4.32\% & 1.75\% & 2.21\% \\
& 32 & 12.37 & 310.78 & 323.15 & 27.81 & 5.58\% & 2.98\% & 2.50\% \\
& 64 & 20.36 & 427.75 & 448.11 & 50.57 & \textbf{2.64\%} & \textbf{0.52\%} & \textbf{1.35\%} \\

\bottomrule
\end{tabular}
}
\end{table*}

As the calibration set grows, the computational cost increases
substantially, while the accuracy improvement is not strictly
monotonic. Using 64 trajectories gives the strongest overall
performance, with 2.64\% Base Loss, 0.52\% Div-Free error, and
1.35\% Vorticity error, but requires 448.11 minutes in total.
In contrast, using only 8 trajectories reduces the total compression
time to 74.55 minutes while retaining competitive performance
(3.55\% Base Loss, 0.83\% Div-Free error, and 2.36\% Vorticity error).
Thus, the calibration size can be selected according to the available
compute budget and desired fidelity; we use 64 trajectories in the
main experiments to prioritize compression quality.

\subsection{Test Dataset}
We evaluate on three benchmarks: Poseidon, VICON, and MPP. For Poseidon, we test all trajectories by considering every ordered snapshot pair $(x_i, x_j)$ within each sequence with $t_i < t_j$, thereby covering both short- and long-range temporal dynamics. For VICON, we use the default test split and evaluate snapshot prediction over all available samples. For MPP, where only a validation dataset is provided, we use the validation split for evaluation and test up to 64 trajectories per sub-dataset. Evaluation spans multiple physical regimes, including incompressible and compressible flows.

\subsection{Evaluation Metrics}
\label{app:detailed_metric}
This section provides detailed definitions of the metrics used to evaluate compression quality. We group the metrics into two categories: (i) model base loss and Sobolev loss, which measure predictive fidelity and derivative consistency, and (ii) physics-specific quantities that assess structural consistency of the solution fields, such as divergence-free error for incompressible Navier--Stokes.

As stated in the main paper, all reported results are expressed as the absolute relative percentage change between the compressed and uncompressed models:

\begin{equation}
\delta_m =
\frac{|m_{\mathrm{comp}}-m_{\mathrm{orig}}|}
{|m_{\mathrm{orig}}|+\epsilon}
\times 100\%,
\end{equation}

where $m_{\mathrm{comp}}$ and $m_{\mathrm{orig}}$ denote the metric values of the compressed and uncompressed models, respectively. The definitions below specify how each metric is computed for an individual model, yielding either $m_{\mathrm{comp}}$ or $m_{\mathrm{orig}}$.

\paragraph{Model Base Loss.}
\label{app:model_base_loss}
The model base loss measures agreement between the model prediction and the ground-truth target under the original training objective.
Let $f_{\theta}(\mathbf{z})$ denote the model prediction for input $\mathbf{z}$, and let $\mathbf{y}$ be the corresponding target.
We evaluate the empirical loss over the test set $\mathcal{D}_{\mathrm{test}}$:
\begin{equation}
\mathcal{L}_{\mathrm{Model}}
=
\frac{1}{|\mathcal{D}_{\mathrm{test}}|}
\sum_{(\mathbf{z},\mathbf{y})\in\mathcal{D}_{\mathrm{test}}}
\ell\!\bigl(f_{\theta}(\mathbf{z}),\mathbf{y}\bigr),
\end{equation}
where $\ell(\cdot,\cdot)$ is the task-specific criterion used during training.
In practice, Poseidon uses relative $l_1$ error, VICON uses mean squared error (MSE), and MPP uses relative MSE.

\paragraph{Sobolev Loss.}
\label{app:sobolev_loss}

To evaluate functional smoothness and derivative consistency across different orders, we adopt the Sobolev loss introduced in Section~\ref{sec:methodology}. While either the $l_1$ or $l_2$ Sobolev loss can be used for sensitivity estimation, we use a fixed $l_1$ Sobolev metric for evaluation, including when evaluating models compressed using the $l_2$ variant.
\begin{equation}
\mathcal{L}_{\mathrm{Sob}}
=
\sum_{|\nu|\le p}
\lambda_{\nu}
\frac{1}{|\Omega_h|}
\sum_{\mathbf{r}\in\Omega_h}
\left\|
D^{\nu}\mathbf{u}(\mathbf r)
-
D^{\nu}\hat{\mathbf{u}}(\mathbf r)
\right\|_1.
\end{equation}
All notation follows the definitions in Section~\ref{sec:methodology}.

\begin{table}[h]
\centering
\small
\caption{Sobolev derivative order $p$ used for evaluation.}
\label{tab:pde_summary_fixed}
\begin{tabularx}{\textwidth}{l c >{\raggedright\arraybackslash}X >{\raggedright\arraybackslash}X >{\raggedright\arraybackslash}X}
\toprule
\textbf{PDE Category} & \textbf{Sobolev Order} & \textbf{Poseidon Datasets} & \textbf{VICON Datasets} & \textbf{MPP Datasets} \\ 
\midrule
Incompressible Navier--Stokes & 2 & NS-PwC, NS-BB & NS--2D & IncompNS \\
\midrule
Compressible Euler & 1 & CE-RM, CE-RPUI & Euler-2D & - \\
\midrule
Compressible Navier--Stokes & 1 & - & Comp-2D & CompressibleNS \\
\midrule
Wave Equations & 1 & Wave-Gauss, Wave-Layer & - & - \\
\midrule
Shallow Water & 1 & - & - & ShallowWater \\
\midrule
Diffusion--Reaction & 2 & - & - & DiffusionReaction \\
\bottomrule
\end{tabularx}
\end{table}

\paragraph{Divergence-Free Error.}
For a predicted two-dimensional velocity field 
$\hat{\mathbf{u}} = (\hat{u}, \hat{v})$, incompressibility requires 
$\nabla \cdot \hat{\mathbf{u}} = 0$.
We quantify the violation of this constraint by the mean absolute divergence:
\begin{equation}
    \mathcal{L}_{\mathrm{div}} 
    =
    \frac{1}{|\Omega_h|}
    \sum_{\mathbf{r} \in \Omega_h} 
    \left| \nabla \cdot \hat{\mathbf{u}}(\mathbf{r}) \right|
    =
    \frac{1}{|\Omega_h|}
    \sum_{\mathbf{r} \in \Omega_h} 
    \left| 
    \frac{\partial \hat{u}}{\partial x}(\mathbf{r})
    +
    \frac{\partial \hat{v}}{\partial y}(\mathbf{r})
    \right|.
    \label{eq:div_loss}
\end{equation}

\paragraph{Vorticity Error.} 
Vorticity measures local rotational dynamics of the flow.
For a two-dimensional velocity field $\mathbf{u}=(u,v)$, the scalar vorticity is defined as
\begin{equation}
    \omega
    =
    \frac{\partial v}{\partial x}
    -
    \frac{\partial u}{\partial y}.
\end{equation}
Similarly, for the predicted velocity field $\hat{\mathbf{u}}=(\hat{u},\hat{v})$, we compute
\begin{equation}
    \hat{\omega}
    =
    \frac{\partial \hat{v}}{\partial x}
    -
    \frac{\partial \hat{u}}{\partial y}.
\end{equation}
We compare the predicted and reference vorticity fields using
\begin{equation}
    \mathcal{L}_{\mathrm{vort}}
    =
    \frac{1}{|\Omega_h|}
    \sum_{\mathbf{r} \in \Omega_h}
    \left|
    \hat{\omega}(\mathbf{r})
    -
    \omega(\mathbf{r})
    \right|.
    \label{eq:vort_loss}
\end{equation}
This metric captures whether compression preserves fine-scale rotational structures that may not be visible from velocity error alone.

Together, these metrics evaluate not only predictive accuracy, but also smoothness, derivative fidelity, and physical consistency of compressed models.

\paragraph{Mass Conservation Error.}
For the compressible Euler equations, the total mass should be conserved.
For the predicted density field $\hat{\rho}$, we compute the spatially
averaged mass at time $t$ as
\begin{equation}
\hat{M}_t
=
\frac{1}{|\Omega_h|}
\sum_{\mathbf{r}\in\Omega_h}
\hat{\rho}_t(\mathbf{r}).
\end{equation}
We quantify the violation of mass conservation by its temporal drift:
\begin{equation}
\mathcal{L}_{\mathrm{mass}}
=
\left|
\hat{M}_t-\hat{M}_{t-\Delta t}
\right|.
\end{equation}

\paragraph{Momentum Conservation Error.}
For a two-dimensional compressible flow, the momentum density is
$\rho\mathbf{v}=\rho(u,v)$. We compute the spatially averaged
predicted momentum as
\begin{equation}
\hat{\mathbf{P}}_t
=
\frac{1}{|\Omega_h|}
\sum_{\mathbf{r}\in\Omega_h}
\hat{\rho}_t(\mathbf{r})
\begin{pmatrix}
\hat{u}_t(\mathbf{r}) \\
\hat{v}_t(\mathbf{r})
\end{pmatrix}.
\end{equation}
The momentum conservation error is then defined as
\begin{equation}
\mathcal{L}_{\mathrm{mom}}
=
\left\|
\hat{\mathbf{P}}_t
-
\hat{\mathbf{P}}_{t-\Delta t}
\right\|_1.
\end{equation}

\paragraph{Total-Energy Conservation Error.}
For the compressible Euler equations, the total-energy density is
\begin{equation}
E(\mathbf{r})
=
\frac{p(\mathbf{r})}{\gamma-1}
+
\frac{1}{2}\rho(\mathbf{r})
\left(
u(\mathbf{r})^2+v(\mathbf{r})^2
\right),
\end{equation}
where $\gamma$ is the ratio of specific heats. Accordingly, the
spatially averaged predicted total energy at time $t$ is
\begin{equation}
\hat{E}_t
=
\frac{1}{|\Omega_h|}
\sum_{\mathbf{r}\in\Omega_h}
\left[
\frac{\hat{p}_t(\mathbf{r})}{\gamma-1}
+
\frac{1}{2}\hat{\rho}_t(\mathbf{r})
\left(
\hat{u}_t(\mathbf{r})^2
+
\hat{v}_t(\mathbf{r})^2
\right)
\right].
\end{equation}
We measure the violation of total-energy conservation as
\begin{equation}
\mathcal{L}_{\mathrm{energy}}
=
\left|
\hat{E}_t-\hat{E}_{t-\Delta t}
\right|.
\end{equation}

\section{Trace-Based Energy Balancing for Multi-Physics Models}
\label{app:trace_balance}

Unified multi-physics models such as VICON~\cite{cao2024vicon} and MPP~\cite{mccabe2024multiple} are calibrated on heterogeneous physical datasets whose state variables can differ greatly in numerical scale. Directly pooling all calibration samples may therefore make the clean input covariance $\boldsymbol{\Sigma}_{\mathbf{XX}}$ and the Fisher sensitivity matrix $\mathbf{F}_{\mathbf Z}=\mathbf{L}^{\top}\mathbf{L}$ dominated by high-energy domains. Since our greedy rank allocation ranks singular components of $\mathbf{L}_l\mathbf{W}_l\mathbf{R}^c_l$, this can bias the allocated ranks toward numerically large domains rather than loss-sensitive ones.

To mitigate this effect, Ours* applies trace-based energy balancing when estimating the calibration statistics used for rank selection. For each layer $l$ and physical dataset $c\in\mathcal{C}$, we compute the category-specific clean covariance $\boldsymbol{\Sigma}^{(c)}_{\mathbf{XX},l}$ and Fisher matrix $\mathbf{F}^{(c)}_{\mathbf Z,l}$. For the input covariance, we define
\begin{equation}
    E^{X}_{l,c}
    =
    \operatorname{Tr}\!\left(
    \boldsymbol{\Sigma}^{(c)}_{\mathbf{XX},l}
    \right),
    \qquad
    \tau^X_l
    =
    \max_{c\in\mathcal{C}} E^{X}_{l,c},
\end{equation}
and rescale each domain by
\begin{equation}
    s^X_{l,c}
    =
    \begin{cases}
    \tau^X_l / E^X_{l,c}, & E^X_{l,c}>\epsilon,\\
    1, & \text{otherwise},
    \end{cases}
\end{equation}
where $\epsilon=10^{-15}$ prevents division by zero. The balanced clean covariance is then
\begin{equation}
    \bar{\boldsymbol{\Sigma}}_{\mathbf{XX},l}
    =
    \frac{1}{|\mathcal{C}|}
    \sum_{c\in\mathcal{C}}
    s^X_{l,c}
    \boldsymbol{\Sigma}^{(c)}_{\mathbf{XX},l}.
\end{equation}
The Fisher matrix is balanced analogously using its own trace:
\begin{equation}
    \bar{\mathbf{F}}_{\mathbf Z,l}
    =
    \frac{1}{|\mathcal{C}|}
    \sum_{c\in\mathcal{C}}
    s^F_{l,c}\mathbf{F}^{(c)}_{\mathbf Z,l},
    \qquad
    s^F_{l,c}
    =
    \frac{
    \max_{c'\in\mathcal{C}}
    \operatorname{Tr}(\mathbf{F}^{(c')}_{\mathbf Z,l})
    }{
    \operatorname{Tr}(\mathbf{F}^{(c)}_{\mathbf Z,l})
    }.
\end{equation}

We then factorize
$\bar{\boldsymbol{\Sigma}}_{\mathbf{XX},l}=\bar{\mathbf{R}}^c_l\bar{\mathbf{R}}^{c\top}_l$
and
$\bar{\mathbf{F}}_{\mathbf Z,l}=\bar{\mathbf{L}}_l^\top\bar{\mathbf{L}}_l$,
and use the singular values of
\begin{equation}
    \bar{\mathbf{L}}_l\mathbf{W}_l\bar{\mathbf{R}}^c_l
\end{equation}
for greedy rank allocation. Thus, the score
\begin{equation}
    v_{l,i}
    =
    \frac{\sigma_{l,i}^2}{P^{(l)}}
\end{equation}
is less affected by raw domain scale and better reflects loss sensitivity across the multi-physics calibration set.

In the implementation, after ranks are selected, we also apply the trace-balancing idea when aggregating the reconstruction statistics in Equation \ref{eq:covariances}. This affects the final weight reconstruction, whereas the rank scores above depend only on the balanced clean covariance and Fisher matrices.
\section{Implementation Details}
\label{app:implementation_details}
\subsection{$\alpha$: Balancing Local Approximation and Error Propagation}

The hyperparameter $\alpha$ regulates the trade-off between preserving the intra layer reconstruction (using clean input $\mathbf{X}$) and maintaining output similarity under actual compression (using shifted input $\mathbf{X'}$), a concept motivated by SAES-SVD \cite{hu2026saes}. Empirical evidence suggests that when input perturbations are minimal, model performance is relatively insensitive to the specific value of $\alpha$. This is because the shifted activation $\mathbf{X'}$ remains similar to the clean input $\mathbf{X}$, allowing our rank selection proxy---which is derived from clean activations---to remain tightly coupled with the weight reconstruction objective. 

However, as the compression ratio increases and perturbations become more pronounced, prioritizing the shifted input (i.e., utilizing a larger $\alpha$) becomes critical. In high-compression regimes, this focus can significantly mitigate performance collapse, for instance, reducing relative loss degradation from $3000\%$ to $1000\%$. Consequently, we set $\alpha = 0.7$ as the default. This provides a robust balance that excels under aggressive compression while remaining competitive at lower compression ratios.

\subsection{$\lambda_\mathrm{Sob}$: Balancing Base Loss and Sobolev Fidelity}

The magnitude of the Sobolev loss often significantly exceeds that of the model base loss, necessitating a rescaling factor $\lambda_\mathrm{Sob}$ to align their numerical ranges. Empirical investigations reveal that once magnitude alignment is achieved, further increasing the weight of the Sobolev loss—specifically to an order of magnitude higher than the base loss—yields superior compression outcomes. This observation suggests that for PFMs, prioritizing high-order structural information provides a potent form of structural regularization. By emphasizing derivative consistency, the framework effectively guides the compression process to preserve governing dynamics and physical laws, which are more sensitive to parameter perturbations than zeroth-order numerical values.

\subsection{Numerical Stability and Positive Definiteness}
\label{app:pd}
Standard whitening techniques typically rely on Cholesky decomposition (e.g. $\mathbf{\Sigma_{c\_cov} = RR^\top}$), which strictly requires the matrix to be positive definite (PD). However, when the number of calibration samples is smaller than the activation dimension (same for $\mathbf{F_Z}$), the covariance matrix becomes rank-deficient and merely positive semi-definite (PSD), causing Cholesky decomposition to fail. 

To overcome this bottleneck, we employ a robust fallback strategy inspired by the dual-SVD formulation in SVD-LLM V2 \cite{wang2025svdv2}. Specifically, we add a negligible diagonal perturbation ($\epsilon I$) to foster PD properties for Cholesky decomposition. If numerical instability persist, we switch to an Eigendecomposition (EVD) approach:
\begin{equation}
    \mathbf{\Sigma_{c\_cov} = U \Lambda U^\top},
\end{equation}

where $\mathbf{U}$ is the orthogonal eigenvector matrix and $\mathbf{\Lambda}$ is the diagonal matrix of non-negative eigenvalues. We then construct the equivalent factor $\mathbf{R = U \Lambda^{1/2}}$. Because EVD is mathematically equivalent to SVD for symmetric PSD matrices, this approach bypasses the strict PD requirement and perfectly preserves the theoretical minimum truncation loss, ensuring highly effective and robust compression even under extreme low-sample regimes.

\subsection{Uniform Rank Allocation in Ablation}
\label{app:uniform_rank}
\paragraph{Uniform Rank Allocation.}
For the uniform-rank-allocation baseline in the ablation study,
we assign the rank of each linear layer according to the same target
parameter retention ratio $r$. For a layer with input and output
dimensions $d_{\mathrm{in}}$ and $d_{\mathrm{out}}$, respectively,
the retained rank is computed as
\begin{equation}
k =
\left\lfloor
\frac{d_{\mathrm{out}} d_{\mathrm{in}}\, r}
     {d_{\mathrm{out}} + d_{\mathrm{in}}}
\right\rfloor .
\end{equation}
A rank-$k$ factorization contains
$k(d_{\mathrm{out}}+d_{\mathrm{in}})$ parameters, so this choice
yields approximately the same parameter retention ratio $r$
for every linear layer.
\section{Baseline Implementation Details}
\label{app:baseline_implementation_detail}
\subsection{ASVD}
We follow the original ASVD pipeline~\cite{yuan2023asvd} without modification to the core algorithm. For activation profiling, we use the magnitude-based scaling method with $\alpha = 0.5$. Sensitivity is estimated via the stable-rank metric, which avoids the cost of forward-pass evaluation. The candidate parameter retention ratios are $\mathcal{R} = \{0.1, 0.2, \ldots, 0.9\}$. Per-layer truncation ranks are determined by binary search under a global Linear-layer parameter budget, and singular values are absorbed evenly into both factors.

\subsection{Dobi-SVD}
We strictly follow the default hyperparameter settings and core differentiable SVD mechanics of the original Dobi-SVD~\cite{wang2025dobi} framework. The only necessary adaptation is a loss scaling mechanism to address the extreme disparity in gradient magnitudes between language and physical models. LLMs optimize Cross-Entropy loss, where the exponentiated loss (Perplexity) typically hovers around $\sim 20$. In contrast, PDE models like VICON optimize a highly normalized Mean Squared Error (MSE) that is orders of magnitude smaller ($\sim 0.001$). Directly applying Dobi-SVD would result in vanishing gradients for the SVD truncation parameters ($\gamma$). To bridge this gap and align the gradient spaces, we introduce an exponential loss scaling strategy: $\mathcal{L}_{scaled} = \exp(w \times \mathcal{L}_\mathrm{Model}) - 1$, with the scaling weight $w$ empirically set to $3000$. This ensures the scaled physical loss yields gradient magnitudes comparable to those in LLMs, allowing the original optimizer logic to function effectively.

\textbf{Discussion.}
Dobi-SVD is vulnerable to a gradient-myopia failure mode when transferred to Poseidon.
Although the exponential loss scaling restores the overall gradient magnitude, it does not alter the locality of the $\tanh$-based truncation derivative. 
Specifically, the soft truncation mask is defined as
\begin{equation}
    T_i(\gamma)
    =
    \frac{1}{2}\tanh\bigl(\beta(\gamma-i)\bigr)
    +
    \frac{1}{2},
\end{equation}
whose derivative with respect to the truncation parameter $\gamma$ is
\begin{equation}
    \frac{\partial T_i}{\partial \gamma}
    =
    \frac{\beta}{2}
    \operatorname{sech}^2\bigl(\beta(\gamma-i)\bigr).
\end{equation}
This derivative is sharply concentrated around the current truncation boundary $i \approx \gamma$, with an effective support controlled by $1/\beta$.
Consequently, a layer initialized with a truncation position far below its true rank requirement receives little or no gradient signal from the higher-rank singular components that would substantially reduce the task loss.

In our implementation, this locality is further amplified because the low-rank SVD is computed only within a small neighborhood of the current $\gamma$.
That is, singular components beyond the current SVD window are absent from the computational graph, rather than merely being assigned small truncation gradients.
As a result, even if increasing the rank of a sensitive layer would significantly improve the physical prediction loss, the optimizer may fail to detect this direction.
Other, less sensitive layers can instead retain higher ranks to satisfy the global compression constraint, leading to a stable but suboptimal rank allocation from which the optimization cannot easily escape.

\subsection{SAES-SVD}
We integrated the SAES-SVD~\cite{hu2026saes} framework strictly following the optimal hyperparameter settings recommended in the original paper. For the Adaptive Collaborative Error Suppression (ACES) mechanism, the authors suggest bounding the alignment strength $\alpha$ within $[0.25, 0.75]$ to achieve the best performance trade-off. We adopted this exact constraint. Based on the framework's formulation $\beta = \frac{\alpha}{1+\alpha}$, this maps to a search interval of $\beta \in [0.2, 0.428]$ in our implementation.

\subsection{FWSVD and SVD-LLM V2}
As FWSVD~\cite{hsu2022language} and SVD-LLM V2~\cite{wang2025svdv2} do not require explicit hyperparameter tuning, we strictly adhere to their official implementation pipelines.
\end{document}